%% file: phrasecut.tex
\documentclass[10pt,twocolumn,letterpaper]{article}

\usepackage{iccv}
\usepackage{times}
\usepackage{epsfig}
\usepackage{graphicx}
\usepackage{amsmath}
\usepackage{amssymb}
\usepackage{mathtools}
\usepackage{subcaption}
\usepackage{float}
\usepackage{booktabs}

\usepackage[english]{babel}
\usepackage[autostyle]{csquotes}
\MakeOuterQuote{"}

\usepackage[pagebackref=true,breaklinks=true,letterpaper=true,colorlinks,bookmarks=false]{hyperref}

\iccvfinalcopy 

\def\iccvPaperID{459} 

\ificcvfinal\fi
\begin{document}

\title{Recurrent Multimodal Interaction for Referring Image Segmentation}

\author{Chenxi Liu$^1$ \quad Zhe Lin$^2$ \quad Xiaohui Shen$^2$ \quad Jimei Yang$^2$ \quad Xin Lu$^2$ \quad Alan Yuille$^1$ \\
Johns Hopkins University$^1$ \quad Adobe Research$^2$ \\
{\tt\small \{cxliu, alan.yuille\}@jhu.edu \quad \{zlin, xshen, jimyang, xinl\}@adobe.com}
}

\maketitle
\thispagestyle{empty}

\input{abstract}


\input{intro}

\input{related}

\input{model}

\input{exp}

\input{conc}

\vspace{-0.4cm}
\paragraph*{Acknowledgments}

We gratefully acknowledge support from NSF CCF-1231216 and a gift from Adobe. 

{\small
\bibliographystyle{ieee}
\bibliography{phrasecut}
}

\end{document}

%% file: abstract.tex
\begin{abstract}
In this paper we are interested in the problem of image segmentation given natural language descriptions, i.e. referring expressions. Existing works tackle this problem by first modeling images and sentences independently and then segment images by combining these two types of representations. We argue that learning word-to-image interaction is more native in the sense of jointly modeling two modalities for the image segmentation task, and we propose convolutional multimodal LSTM to encode the sequential interactions between individual words, visual information, and spatial information. We show that our proposed model outperforms the baseline model on benchmark datasets. In addition, we analyze the intermediate output of the proposed multimodal LSTM approach and empirically explain how this approach enforces a more effective word-to-image interaction.\footnote{Code is available at \url{https://github.com/chenxi116/TF-phrasecut-public}}
\end{abstract}


%% file: intro.tex
\section{Introduction}



In this paper, we study the challenging problem of using natural language expressions to segment an image. 
Given both an image and a natural language expression, we are interested in segmenting out the corresponding region referred by the expression. 
This problem was only introduced recently, but has great value as it provides new means for interactive image segmentation. 
Specifically, people can segment/select image regions of their interest by typing natural language descriptions or even speaking to the computer \cite{DBLP:conf/chi/LaputDWCALA13}. 

Given the success of convolutional neural networks in semantic segmentation \cite{DBLP:conf/cvpr/LongSD15, DBLP:journals/corr/ChenPKMY14, DBLP:journals/corr/ChenPK0Y16}, an immediate way to tackle this problem is to augment the convolutional semantic segmentation networks with a LSTM \cite{DBLP:journals/neco/HochreiterS97} sentence encoder \cite{DBLP:conf/eccv/HuRD16}, so that the image features and sentence representation can be combined to produce the desired mask.
In fact, this sentence-to-image interaction scheme has been also adopted by recent methods on referring object localization \cite{DBLP:conf/cvpr/ZhuGBF16} and visual question answering tasks \cite{DBLP:conf/iccv/AntolALMBZP15}. 

\begin{figure}[t]
\captionsetup[subfigure]{labelformat=empty, font=small}
\centering
\begin{subfigure}{0.46\textwidth}
\centering
\begin{subfigure}{0.32\textwidth}
\caption{\textit{$\hdots$standing}}
\vspace{-2mm}
\includegraphics[width=\linewidth]{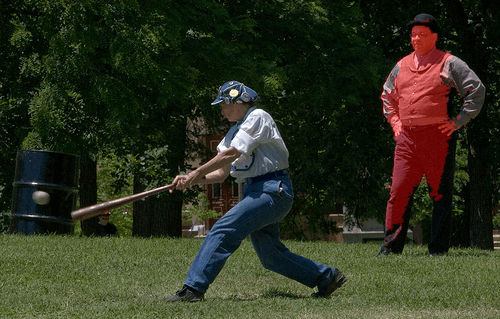}
\includegraphics[width=\linewidth]{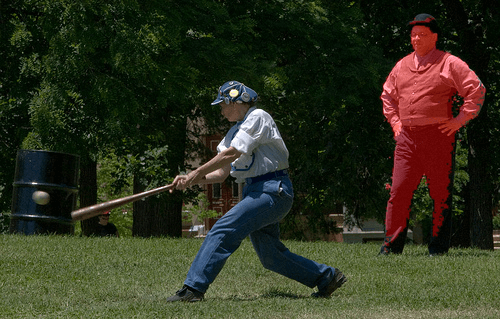}
\end{subfigure}
\begin{subfigure}{0.32\textwidth}
\caption{\textit{$\hdots$someone}}
\vspace{-2mm}
\includegraphics[width=\linewidth]{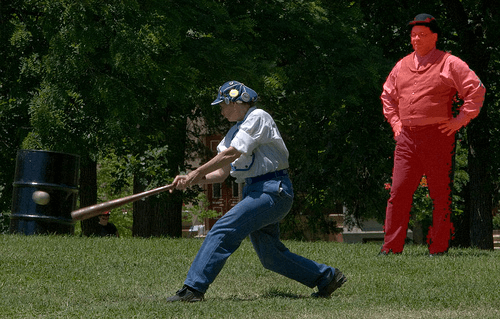}
\includegraphics[width=\linewidth]{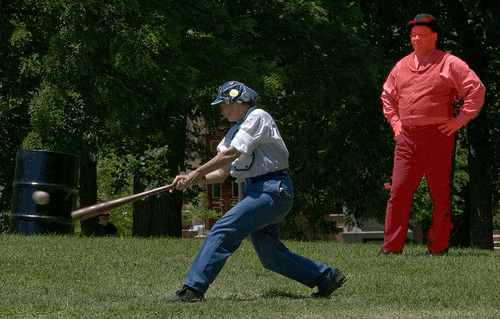}
\end{subfigure}
\begin{subfigure}{0.32\textwidth}
\caption{\textit{$\hdots$bat}}
\vspace{-2mm}
\includegraphics[width=\linewidth]{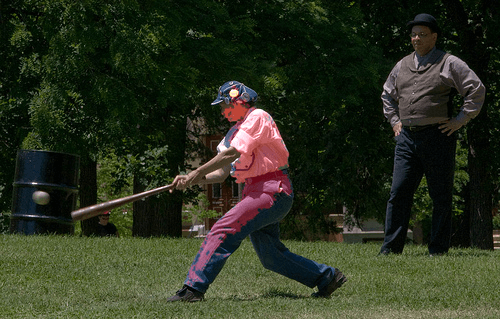}
\includegraphics[width=\linewidth]{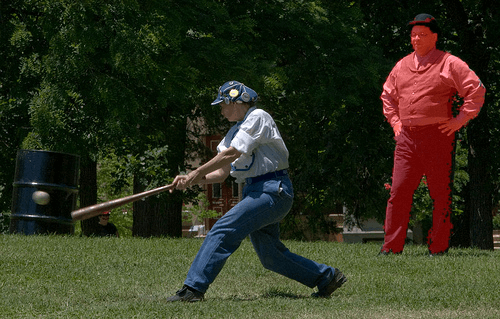}
\end{subfigure}
\caption{\textit{Man in a vest and blue jeans \underline{standing} watching \underline{someone} swing a \underline{bat}. }}
\end{subfigure}
\caption{Given the image and the referring expression, we are interested in segmenting out the referred region. 
Each column shows segmentation result until after reading the underlined word.
Our model (second row) explicitly learns the progression of multimodal interaction with convolutional LSTM, which helps long-term memorization and correctly segments out the referred region compared with the baseline model (first row) which uses language-only LSTM. }
\end{figure}

However, this sentence-to-image scheme does not reflect how humans tackle this problem.
In sentence-picture verification, it is found through eye tracking that when pictures and sentences are presented together, people either follow a image-sentence-image reading sequence, or go back-and-forth between sentence and picture a number of times before making the decision \cite{underwood2004inspecting}.
In other words, the interaction between image and sentence should prevail from the beginning to the end of the sentence, instead of only happening at the end of the sentence.
Presumably this is because the semantic information is more concrete and therefore more easily remembered when grounded onto the image. 
For example, consider the expression "the man on the right wearing blue".
Without seeing an actual image, all information in the sentence needs to be remembered, meaning the sentence embedding needs to encode \texttt{IS\_MAN, ON\_RIGHT, WEAR\_BLUE} jointly. 
However, with the actual image available, the reasoning process can be decomposed as a sequential process, where the model first identifies all pixels that agree with \texttt{IS\_MAN}, then prunes out those that do not correspond with \texttt{ON\_RIGHT}, and finally suppresses those that do not agree with \texttt{WEAR\_BLUE}.


Motivated by this sequential decision making theory, we propose a two-layered convolutional multimodal LSTM network that explicitly models word-to-image interaction.
Different from the language-only LSTM encoder in previous works \cite{DBLP:conf/eccv/HuRD16}, the convolutional multimodal LSTM takes both visual feature and language representation as input to generate the hidden state that retains both the spatial and semantic information in memory.
Therefore its hidden state models how the multimodal feature progresses over time  or word-reading order.
After seeing the last word, we use a convolution layer to generate the image segmentation result.

In summary, the contribution of our paper is three-fold:
\begin{itemize}
\item We propose a novel model, namely convolutional multimodal LSTM, to encode the sequential interactions between individual semantic, visual, and spatial information. 
\item We demonstrate the superior performance of the word-to-image multimodal LSTM approach on benchmark datasets over the baseline model.
\item We analyze the intermediate output of the proposed multimodal LSTM approach and empirically explain how this approach enforces a more effective word-to-image interaction. 
\end{itemize}

%% file: related.tex
\section{Related Work}

In this section, we review recent studies that are tightly related to our work in the following three areas: semantic segmentation, referring expression localization, and multimodal interaction representation.

\noindent
\textbf{Semantic Segmentation}
Many state-of-the-art semantic segmentation models employ a fully convolutional network \cite{DBLP:conf/cvpr/LongSD15} architecture. 
FCN converts the fully connected layers in VGG network \cite{DBLP:journals/corr/SimonyanZ14a} into convolutional layers, thereby allowing dense (although downsampled) per-pixel labeling.
However, too much downsampling (caused by pooling layers in the VGG architecture) prohibits the network from generating high quality segmentation results.
DeepLab \cite{DBLP:journals/corr/ChenPKMY14} alleviates this issue by discarding two pooling operations with atrous convolution.
With Residual network \cite{DBLP:conf/cvpr/HeZRS16} as its backbone architecture, DeepLab \cite{DBLP:journals/corr/ChenPK0Y16} is one of the leading models on Pascal VOC \cite{DBLP:journals/ijcv/EveringhamGWWZ10}. 
We use both ResNet-101 (with atrous convolution) and DeepLab ResNet-101 to extract image features in a fully convolutional manner.
Following \cite{DBLP:journals/corr/ChenPKMY14, DBLP:journals/corr/ChenPK0Y16}, we also report the result of using DenseCRF \cite{DBLP:conf/nips/KrahenbuhlK11} for refinement.


\noindent
\textbf{Referring Expression Localization} 
More and more interest arise recently in the problem of localizing objects based on a natural language expression. 
In \cite{DBLP:conf/cvpr/MaoHTCY016} and \cite{DBLP:conf/cvpr/HuXRFSD16}, image captioning models \cite{DBLP:journals/corr/MaoXYWY14a, DBLP:conf/cvpr/DonahueHGRVDS15} are modified to score the region proposals, and the one with the highest score is considered as the localization result.
In \cite{DBLP:conf/eccv/RohrbachRHDS16}, the alignment between the description and image region is learned by reconstruction with attention mechanism.
\cite{DBLP:conf/eccv/YuPYBB16} improved upon \cite{DBLP:conf/cvpr/MaoHTCY016} by explicitly handling objects of the same class within the same image, while \cite{DBLP:conf/eccv/NagarajaMD16} focused on discovering interactions between the object and its context using multiple-instance learning. 
However all these works aim at finding a bounding box of the target object instead of segmentation mask.
Perhaps the most relevant work to ours is \cite{DBLP:conf/eccv/HuRD16}, which studies the same problem of image segmentation based on referring expressions. 
Our approach differs in that we model the \textit{sequential} property of interaction between natural language, visual, and spatial information.
In particular, we update the segmentation belief after seeing each word. 

\noindent
\textbf{Multimodal Interaction Representation} 
Our work is also related to multimodal feature fusion in visual question answering \cite{DBLP:conf/nips/KimLKHKHZ16, DBLP:conf/emnlp/FukuiPYRDR16, DBLP:conf/iccv/MalinowskiRF15} and image captioning \cite{DBLP:conf/cvpr/DonahueHGRVDS15}. 
In \cite{DBLP:conf/cvpr/DonahueHGRVDS15} the input to LSTM is the image feature and the previous word's embedding, whereas in \cite{DBLP:conf/iccv/MalinowskiRF15} the input to LSTM is the image feature and individual question word's embedding.
Attention mechanism \cite{DBLP:conf/cvpr/YangHGDS16, DBLP:conf/icml/XuBKCCSZB15, DBLP:conf/nips/YangYWCS16, DBLP:conf/aaai/LiuMSY17, DBLP:journals/corr/LiuSWWY17} may also be applied, mostly to improve the relevance of image features. 
In both tasks the goal is to generate a textual sequence.
Here instead, we use the LSTM hidden states to generate segmentation, which is not commonly considered a sequential task and requires preservation of spatial location.
We achieve this by applying LSTM in a convolutional manner \cite{DBLP:conf/nips/ShiCWYWW15, DBLP:conf/eccv/ChoyXGCS16, DBLP:conf/nips/FinnGL16}, unlike prior work on recurrent attention \cite{DBLP:conf/nips/MnihHGK14, DBLP:conf/cvpr/KuenWW16}. 

%% file: model.tex
\section{Models}

In this section, we first introduce our notation for this problem (section \ref{sec: notation}), and then describe the baseline model based on the sentence-to-image scheme \cite{DBLP:conf/eccv/HuRD16} (section \ref{sec: baseline}), which only models the progression of semantics.
In section \ref{sec: mlstm} we propose convolutional multimodal LSTM for fusing both modalities and model the progression of multimodal features in addition to the progression of semantics.

\subsection{Notation}
\label{sec: notation}

In the referring image segmentation problem, we are given both an image $I$ and a natural language description $S = \{ w_1, w_2, \hdots, w_T\}$, where $w_t$ ($t \in \{1, 2, \hdots, T\}$) are individual words in the sentence. 
The goal is to segment out the corresponding region in the image. 
We will use $R$ for prediction and $\hat{R}$ for ground truth. 
$R^{ij} \in (0, 1)$ represents the foreground probability of a pixel, where $i$ and $j$ are spatial coordinates.
$\hat{R}^{ij} \in \{0, 1\}$, where 1 means the pixel is referred to by $S$ and 0 otherwise.

\subsection{Baseline Model}
\label{sec: baseline}

\begin{figure}[t]
\centering
\includegraphics[width=0.99\linewidth, trim = {260, 30, 170, 60}, clip]{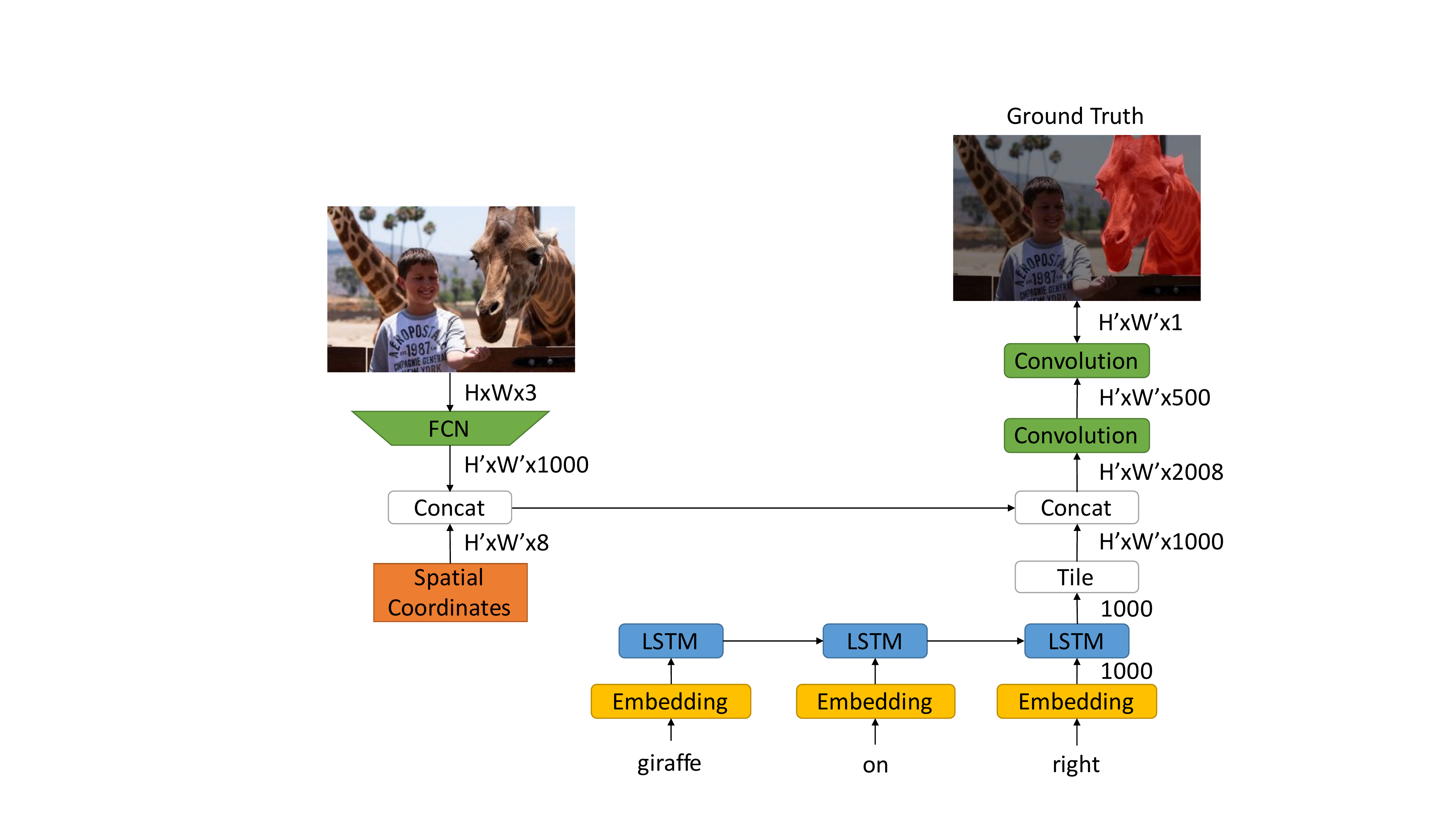}
\caption{Network architecture of the baseline model described in section \ref{sec: baseline}. 
In this model, the entire sentence is encoded into a fixed vector with language-only LSTM without using visual information.
}
\vspace{-0.2cm}
\label{fig: model-baseline}
\end{figure}

Our model is based on the model proposed in \cite{DBLP:conf/eccv/HuRD16}. 
In \cite{DBLP:conf/eccv/HuRD16}, given an image of size $W \times H$, an FCN-32s \cite{DBLP:conf/cvpr/LongSD15} is used to extract image features with size $W' \times H' \times D_I$, where $W' = W / 32$ and $H' = H / 32$. 
The image features are then concatenated with spatial coordinates to produce a $W' \times H' \times (D_I + 8)$ tensor. 
The 8 spatial coordinate dimensions follow the implementation of \cite{DBLP:conf/eccv/HuRD16}. 
The normalized horizontal/vertical position uses 3 dimensions each. 
The remaining 2 dimensions are $1/W'$ and $1/H'$.
We use $\mathbf{v}^{ij} \in \mathbb{R}^{D_I + 8}$ to represent the image-spatial feature at a specific spatial location. 

As for the referring expression, every word $w_t$ is one-hot encoded and mapped to a word embedding $\mathbf{w}_t$. 
The entire sentence is then encoded with an LSTM into a vector $\mathbf{h}_T$ of size $D_S$, where $\mathbf{h}_t$ represents the hidden state of LSTM at time step $t$:
\begin{align}
&\text{LSTM}: (\mathbf{w}_t, \mathbf{h}_{t-1}, \mathbf{c}_{t-1}) \rightarrow (\mathbf{h}_t, \mathbf{c}_t) \label{eqn: slstm} \\
&\begin{pmatrix}
\mathbf{i} \\
\mathbf{f} \\
\mathbf{o} \\
\mathbf{g}
\end{pmatrix}
= \begin{pmatrix}
\text{sigm} \\
\text{sigm} \\
\text{sigm} \\
\text{tanh}
\end{pmatrix}
M_{4n, D_S + n} 
\begin{pmatrix}
\mathbf{w}_t \\
\mathbf{h}_{t-1}
\end{pmatrix} \\
& \mathbf{c}_t = \mathbf{f} \odot \mathbf{c}_{t-1} + \mathbf{i} \odot \mathbf{g} \\
& \mathbf{h}_t = \mathbf{o} \odot \text{tanh}(\mathbf{c}_t)
\end{align}
where $n$ is the size of the LSTM cell.
$\mathbf{i}, \mathbf{f}, \mathbf{o}, \mathbf{g}$ are the input gates, forget gates, output gets, and memory gates respectively. 
$\mathbf{c}_t$ are the memory states at time step $t$. 

The vector $\mathbf{h}_T$ is then concatenated with the image features and spatial coordinates at all locations to produce a $W' \times H' \times (D_I + D_S + 8)$ tensor. 
Two additional convolutional layers and one deconvolution layer are attached to the tensor to produce the final segmentation mask $R \in \mathbb{R}^{W \times H}$.

Given the ground truth binary segmentation mask $\hat{R}$, the loss function is 
\begin{align}
L_{high} = \frac{1}{WH} &\sum_{i=1}^W \sum_{j = 1}^H \Big( \hat{R}^{ij} *-\log(R^{ij}) \nonumber \\
&+ (1 - \hat{R}^{ij}) *-\log(1 - R^{ij}) \Big)  \label{eqn: losshigh}
\end{align}
The whole network is trained with standard back-propagation. 

Our baseline employs the same architecture, except that we use ResNet-101 \cite{DBLP:conf/cvpr/HeZRS16} instead of FCN-32s to extract image features. 
One limitation of FCN-32s is that downsampling by 32 makes $W'$ and $H'$ too small.
Therefore similar to the treatment of DeepLab \cite{DBLP:journals/corr/ChenPKMY14, DBLP:journals/corr/ChenPK0Y16}, we reduce the stride of conv4\_1 and conv5\_1 in ResNet-101 from 2 to 1, and use atrous convolution of rate 2 and 4 to compensate for the change.
This operation reduces the downsampling rate from 32 to 8, which is relatively dense and allows loss to be computed at the feature resolution ($W' = W/8, H' = H/8$) instead of the image resolution. 
 Therefore in our model, the loss function becomes
 \begin{align}
 L_{low} = \frac{1}{W' H'} &\sum_{i=1}^{W'} \sum_{j = 1}^{H'} \Big( \hat{R}^{ij} *-\log(R^{ij}) \nonumber \\
 &+ (1 - \hat{R}^{ij}) *-\log(1 - R^{ij}) \Big) \label{eqn: losslow}
 \end{align}
We use bilinear interpolation to upsample $R \in \mathbb{R}^{W' \times H'}$ at test time. 

We are going to show in the experimental section that combining ResNet with atrous convolution results in a more competitive baseline model and easier training procedure.

\subsection{Recurrent Multimodal Interaction Model}
\label{sec: mlstm}

In the baseline model described above, segmentation is performed once, after the model has seen and memorized the entire referring expression.
The memorization is the process of updating LSTM hidden states while scanning the words in the expression one by one. 
However, as discussed earlier, this requires the model to memorize all the attributes in the sentence jointly.
We instead utilize the sequential property of natural language and turn referring image segmentation into a sequential process.
This requires the language model to have access to the image from the beginning of the expression, allowing the semantics to be grounded onto the image early on.
Therefore we consider modeling of the multimodal interaction, i.e. a scheme that can memorize the \textit{multimodal} information (language, image, spatial information, and their interaction), which has direct influence on the segmentation prediction.

\begin{figure*}[t]
\centering
\includegraphics[width=0.6\linewidth, trim = {120, 55, 350, 80}, clip]{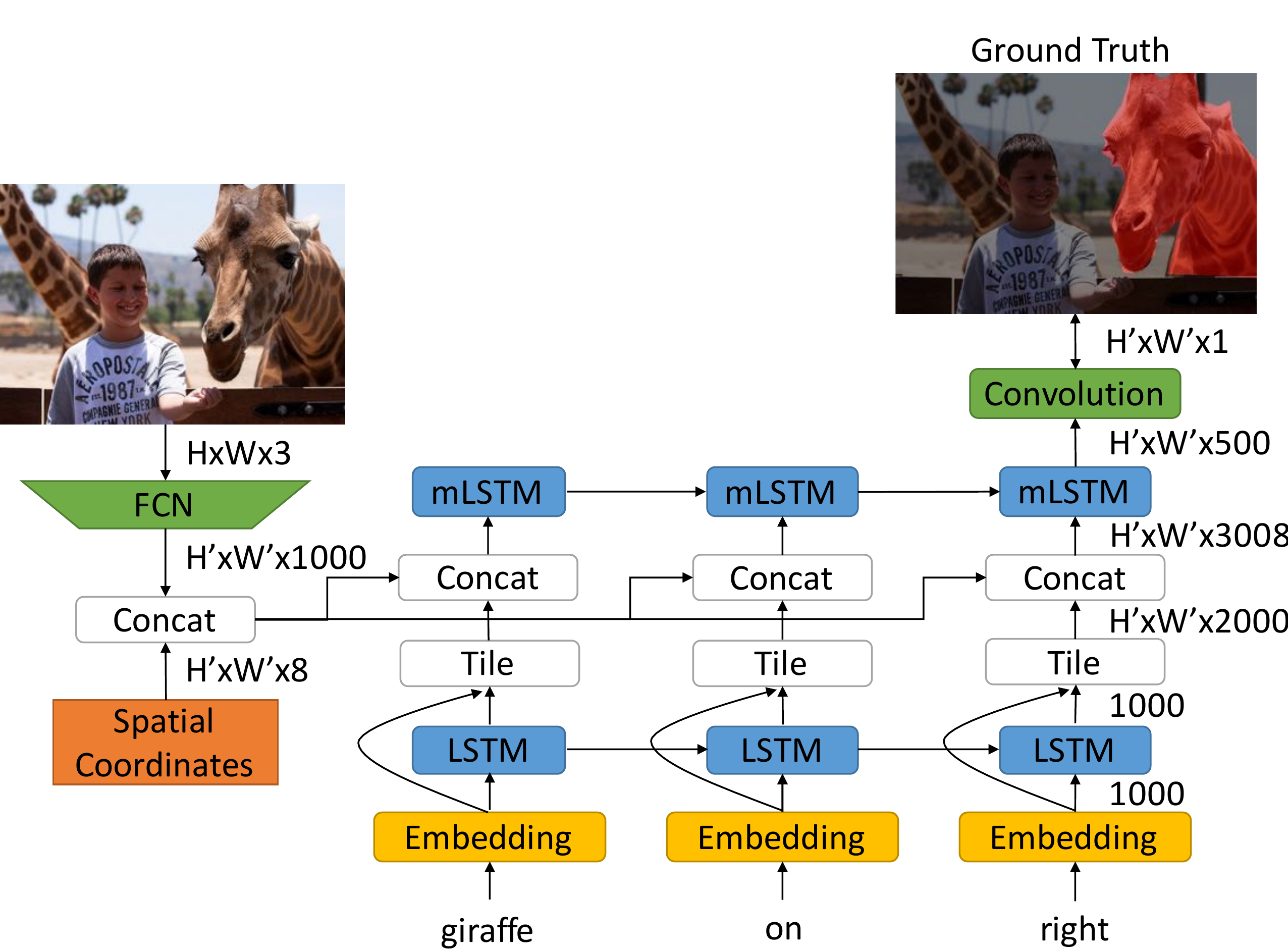}
\caption{Network architecture of the RMI model described in section \ref{sec: mlstm}.
By using the convolutional multimodal LSTM, our model allows multimodal interaction between language, image, and spatial information at each word. 
The mLSTM is applied to all location in the image and implemented as a $1 \times 1$ convolution.
}
\vspace{-0.2cm}
\label{fig: mlstm}
\end{figure*}

We use a multimodal LSTM to capture the progression of rich multimodal information through time as shown in Fig.~\ref{fig: mlstm}. 
Specifically, a multimodal LSTM (mLSTM) uses the concatenation of the language representation $\mathbf{l}_t \in \mathbb{R}^{D_S}$ and the visual feature at a specific spatial location $\mathbf{v}_{ij} \in \mathbb{R}^{D_I + 8}$ as its input vector:
\begin{equation}
\text{mLSTM}: (\begin{bmatrix}
\mathbf{l}_t \\
\mathbf{v}^{ij}
\end{bmatrix}, \mathbf{h}_{t-1}^{ij}, \mathbf{c}_{t-1}^{ij}) \rightarrow (\mathbf{h}_t^{ij}, \mathbf{c}_t^{ij}) \label{eqn: mlstm}
\end{equation}

The same mLSTM operation is shared for all image locations.
This is equivalent to treating the mLSTM as a $1 \times 1$ convolution over the feature map of size $W' \times H' \times (D_I + D_S + 8)$. 
In other words, this is a convolutional LSTM that shares weights both across spatial location and time step.

The baseline model uses language-only LSTM (Equation \ref{eqn: slstm}) to encode the referring expression, and concatenate it with the visual feature to produce $\begin{bmatrix}
\mathbf{h}_T \\
\mathbf{v}^{ij}
\end{bmatrix}$.
One advantage of multimodal LSTM is that either of the two components can be produced by it.
The matrix $M$ in multimodal LSTM will be of size $4n \times (D_S + D_I + 8 + n)$. 
If $M_{1:4n, D_S + 1: D_S + D_I + 8} = \mathbf{0}$, then the mLSTM will essentially ignore the visual part of the input, and encode only the semantic information.
On the other hand, if the mLSTM ignores the language representation, the mLSTM will see the same input $\mathbf{v}_{ij}$ at all time steps, therefore very likely to retain that information.

From another perspective, multimodal LSTM forces word-visual interaction and generates multimodal feature at every recurrent step, which is key to good segmentation.
In the baseline model, in order for the language representation to reach the multimodal level, it has to go through all subsequent LSTM cells as well as a convolution layer:
\begin{equation}
\mathbf{l}_t \xrightarrow{LSTM} \mathbf{h}_T \xrightarrow{Concat} \begin{bmatrix}
\mathbf{h}_T \\
\mathbf{v}^{ij}
\end{bmatrix}
\xrightarrow{Conv} \text{multimodal feature}
\end{equation}
while with multimodal LSTM this can be done with just the (multimodal) LSTM cells:
\begin{equation}
\mathbf{l}_t \xrightarrow{Concat} \begin{bmatrix}
\mathbf{l}_t \\
\mathbf{v}^{ij}
\end{bmatrix}
\xrightarrow{mLSTM} \text{multimodal feature}
\end{equation}
Note that the visual feature still only needs one weight layer to become multimodal.

In our Recurrent Multimodal Interaction (RMI) model, we take the language representation $\mathbf{l}_t$ to be the concatenation of language-only LSTM hidden state in Equation \ref{eqn: slstm} and word embedding $\begin{bmatrix}
\mathbf{h}_t \\ 
\mathbf{w}_t
\end{bmatrix}$.  
This forms a two-layer LSTM structure, where the lower LSTM only encodes the semantic information, while the upper LSTM generates the multimodal feature.
The lower language-only LSTM is spatial-agnostic, while the upper multimodal LSTM preserves feature resolution $H' \times W'$.

%% file: exp.tex
\section{Experiments}

\begin{table*}[t]
\caption{Comparison of segmentation performance (IOU). In the first column, R means ResNet weights, D means DeepLab weights, and DCRF means DenseCRF.}
\centering
\begin{tabular}{c || c | ccc | ccc | c}
\toprule
& Google-Ref & \multicolumn{3}{|c|}{UNC} & \multicolumn{3}{|c|}{UNC+} & ReferItGame \\
& val & val & testA & testB & val & testA & testB & test \\
\midrule
\midrule
\cite{DBLP:conf/eccv/HuRD16, DBLP:journals/corr/HuRVD16} & 28.14 & - & - & - & - & - & - & 48.03 \\
\midrule
\midrule
R+LSTM & 28.60 & 38.74 & 39.18 & 39.01 & 26.25 & 26.95 & 24.57 & 54.01 \\
R+RMI & \bf{32.06} & \bf{39.74} & \bf{39.99} & \bf{40.44} & \bf{27.85} & \bf{28.69} & \bf{26.65} & \bf{54.55} \\
\midrule
R+LSTM+DCRF & 28.94 & 39.88 & 40.44 & 40.07 & 26.29 & 27.03 & 24.44 & 55.90 \\
R+RMI+DCRF & \bf{32.85} & \bf{41.17} & \bf{41.35} & \bf{41.87} & \bf{28.26} & \bf{29.16} & \bf{26.86} & \bf{56.61} \\
\midrule
D+LSTM & 33.08 & 43.27 & 43.60 & 43.31 & 28.42 & 28.57 & 27.70 & 56.83 \\
D+RMI & \bf{34.40} & \bf{44.33} & \bf{44.74} & \bf{44.63} & \bf{29.91} & \bf{30.37} & \bf{29.43} & \bf{57.34} \\
\midrule
D+LSTM+DCRF & 33.11 & 43.97 & 44.25 & 44.07 & 28.07 & 28.29 & 27.44 & 58.20 \\
D+RMI+DCRF & \bf{34.52} & \bf{45.18} & \bf{45.69} & \bf{45.57} & \bf{29.86} & \bf{30.48} & \bf{29.50} & \bf{58.73} \\
\bottomrule
\end{tabular}
\label{tab: iou}
\end{table*}

\subsection{Datasets}

We use four datasets to evaluate our model: Google-Ref \cite{DBLP:conf/cvpr/MaoHTCY016}, UNC \cite{DBLP:conf/eccv/YuPYBB16}, UNC+ \cite{DBLP:conf/eccv/YuPYBB16}, and ReferItGame \cite{DBLP:conf/emnlp/KazemzadehOMB14}.

Google-Ref contains 104560 expressions referring to 54822 objects from 26711 images selected from MS COCO \cite{DBLP:conf/eccv/LinMBHPRDZ14}. 
These images all contain 2 to 4 objects of the same type.
In general the expressions are longer and with richer descriptions, with an average length of 8.43 words.
Although the dataset has primarily been used for referring object detection \cite{DBLP:conf/cvpr/MaoHTCY016, DBLP:conf/eccv/YuPYBB16, DBLP:conf/eccv/NagarajaMD16}, where the goal is to return a bounding box of the referred object, it is also suitable for referring image segmentation, since the original MS COCO annotation contains segmentation masks.
We use the same data split as \cite{DBLP:conf/cvpr/MaoHTCY016}.

UNC and UNC+ are also based on MS COCO images.
Different from Google-Ref, these two datasets are collected interactively in a two-player game \cite{DBLP:conf/emnlp/KazemzadehOMB14}.
The difference between the two datasets is in UNC no restrictions are enforced on the referring expression, while in UNC+ no location words are allowed in the expression, meaning the annotator has to describe the object purely by its appearance.
UNC consists of 142209 referring expressions for 50000 objects in 19994 images, and UNC+ consists of 141564 expressions for 49856 objects in 19992 images.
We use the same data split as \cite{DBLP:conf/eccv/YuPYBB16}.

ReferItGame contains 130525 expressions referring to 96654 distinct objects in 19894 natural images.
Different from the other three datasets, ReferItGame contains "stuff" segmentation masks, such as "sky" and "water", in addition to objects. 
In general the expressions are shorter and more concise, probably due to the collection process as a two-player game. 
We use the same data split as \cite{DBLP:conf/eccv/HuRD16}.

\subsection{Implementation Details}

\cite{DBLP:conf/eccv/HuRD16, DBLP:journals/corr/HuRVD16} both use the VGG network \cite{DBLP:journals/corr/SimonyanZ14a} pretrained on ImageNet \cite{DBLP:conf/cvpr/DengDSLL009} as visual feature extractor.
We instead experiment with two alternatives: ResNet-101 pretrained on ImageNet, and DeepLab-101 finetuned on Pascal VOC \cite{DBLP:journals/ijcv/EveringhamGWWZ10}. 

In our experiments, we resize (while keeping aspect ratio) and pad (with zero) all images and ground truth segmentation to $W \times H$, and in all our experiments $W = H = 320$.
As for the feature resolution $W' = H' = 40$.
The image feature has dimension $D_I = 1000$, and the sentence vector has dimension $D_S = 1000$.
We choose the cell size of mLSTM to be 500. 
For referring expressions of length more than 20, we only keep the first 20 words. 
All architecture details are in Fig. \ref{fig: model-baseline} \ref{fig: mlstm}, where sizes of blobs are marked.

In \cite{DBLP:conf/eccv/HuRD16} a three-stage training strategy is used.
A detection network is first trained, which is used to initialize the low resolution version of the model. 
After training the low resolution version with $W' = H' = 16$, it is again used to initialize the high resolution version, where a deconvolution layer is learned. 
We instead only train once using the loss function defined in Equation \ref{eqn: losslow}, and observe fast convergence. 
This is probably due to the higher spatial resolution allowed by atrous convolution. 
We use the Adam \cite{DBLP:journals/corr/KingmaB14} optimizer with a fixed learning rate of 0.00025. 
We set the batch size to 1 and weight decay to 0.0005. 

We evaluate using two metrics: Precision@X ($X \in \{0.5, 0.6, 0.7, 0.8, 0.9\}$) and Intersection-over-Union (IOU), where Precision@X means the percentage of images with IOU higher than X. 
This is consistent with previous work \cite{DBLP:conf/eccv/HuRD16, DBLP:journals/corr/HuRVD16} to allow for comparison.
We report the most indicative IOU in the main paper, and the full tables containing Precision numbers are in the supplementary material.

In addition to evaluating the direct segmentation output, we also report results after applying DenseCRF \cite{DBLP:conf/nips/KrahenbuhlK11} for refinement.
We use the same hyperparameters used in \cite{DBLP:journals/corr/ChenPK0Y16}.

\begin{figure}[t]
\centering
\includegraphics[width=0.7\linewidth, trim={0, 180, 0, 210}, clip]{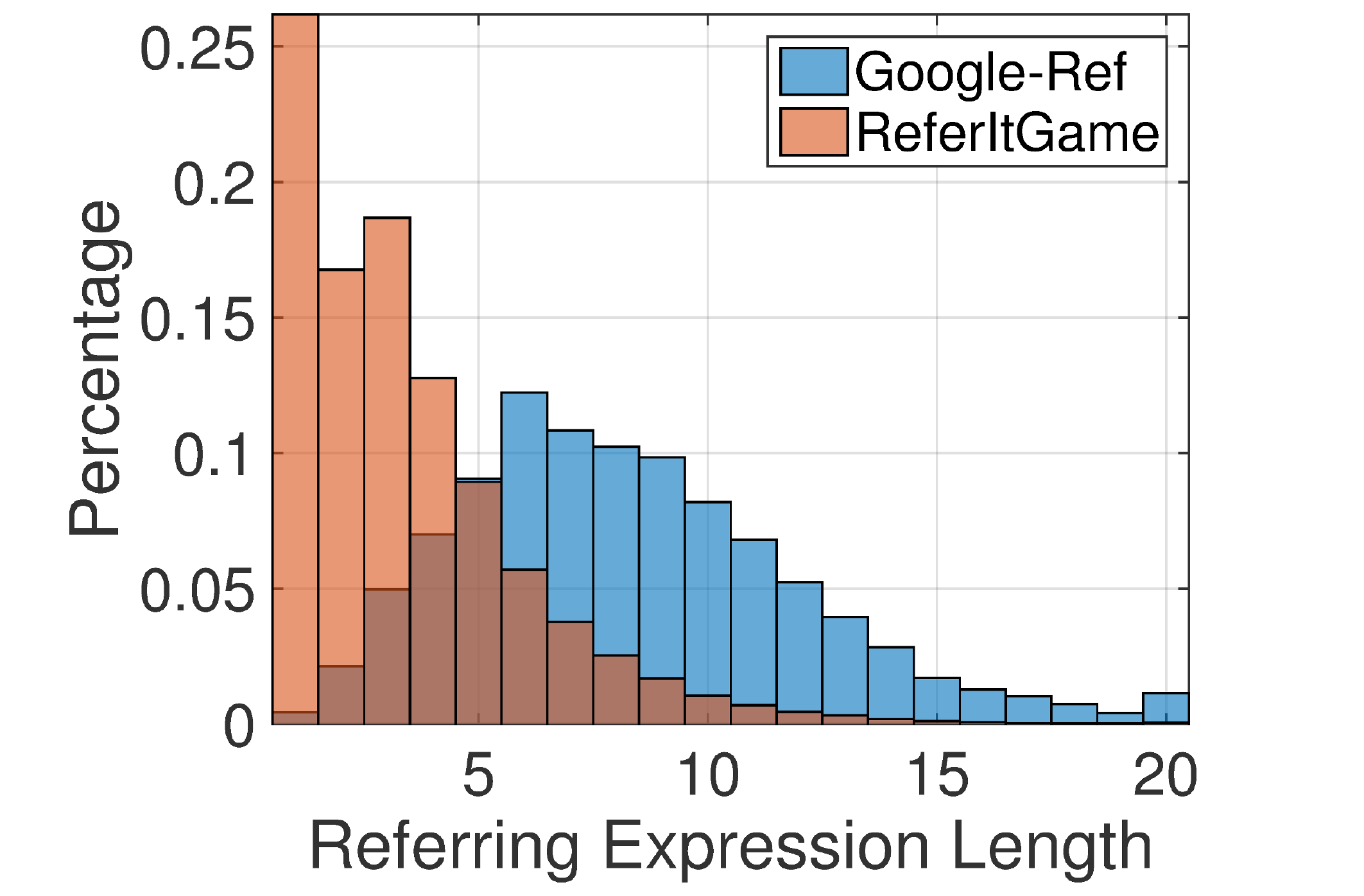}
\vspace{-0.2cm}
\caption{The distribution of referring expression length in the Google-Ref and ReferItGame test set. 
Most of the referring expressions in ReferItGame are short, with over 25 percent single word description.
The distributions of UNC and UNC+ are very similar to that of ReferItGame since the data collection method is the same.
}
\vspace{-0.2cm}
\label{fig: testset}
\end{figure}

\subsection{Quantitative Results}

The segmentation performance (IOU) on all datasets are summarized in Table \ref{tab: iou}.

We first observe that the performance consistently increases by replacing the VGG-based FCN-32s with ResNet. 
This indicates that ResNet can provide better image features for segmentation purpose, which likely comes from both stronger network and higher spatial resolution.
DeepLab delivers even higher baseline since its weights have been finetuned on segmentation datasets, which makes the knowledge transfer easier.

We then study the effect of mLSTM.
Our RMI models with mLSTM consistently outperform those with language-only LSTM by a large margin regardless of the image feature extractor and dataset. 
This shows that mLSTM can successfully generate multimodal features that improve segmentation.
Specifically, on the Google-Ref dataset using ResNet weights, we observe an IOU increase of nearly 3.5\% over the baseline model.

By comparison, the performance increase using mLSTM is not as high on ReferItGame. 
One reason is that the dataset is easier as indicated by the metrics (over 20 percent higher IOU than Google-Ref), and the baseline model already performs well.
Another reason is that the descriptions in this dataset are in general much shorter (see Fig. \ref{fig: testset}), and as a result sequential modeling does not have as much effect.
In fact, over 25 percent images in the ReferItGame test set only has one word as its description.

Another interesting observation is that the performance is considerably worse on UNC+ than on UNC (over 10 percent IOU difference).
As aforementioned, the only difference between the two datasets is in UNC+ there is no spatial/location indicator words that the model can utilize, and the model must understand the semantics in order to output correct segmentation.
This suggests that the LSTM language encoder may be the main barrier in referring image segmentation performance. 

We further show the advantage of our mLSTM model in sequential modeling by breaking down the IOU performance.
Specifically, we want to study the relationship between IOU and referring expression length.
To this end, we split the test set into 4 groups of increasing referring expression length with roughly equal size, and report the individual IOU on these groups.
The results are summarized in Table \ref{tab: gref-iou} \ref{tab: unc-iou} \ref{tab: unc+-iou} \ref{tab: referit-iou}.
Our RMI model outperforms the baseline model in every group.
More interestingly, the relative gain of our RMI model over the baseline model in general increases with the length of the referring expression.
This suggests that mLSTM is better at fusing features over longer sequences, which we will also verify visually.

Finally, by applying the DenseCRF, we observe consistent improvement in terms of IOU.
In addition, the IOU improvement on our RMI model is usually greater than the IOU improvement on the baseline model, suggesting that our model has better localization ability.

\begin{table}[t]
\caption{IOU performance break-down on Google-Ref. }
\centering
\vspace{-0.2cm}
\begin{tabular}{c c c c c}
\toprule
\bf{Length} & \bf{1-5} & \bf{6-7} & \bf{8-10} & \bf{11-20} \\
\midrule
R + LSTM & 32.29 & 28.27 & 27.33 & 26.61 \\
R + RMI & 35.34 & 31.76 & 30.66 & 30.56 \\
Relative Gain & 9.44\% & 12.37\% & 12.17\% & 14.81\% \\
\bottomrule
\end{tabular}
\vspace{-0.2cm}
\label{tab: gref-iou}
\end{table}

\begin{table}[t]
\caption{IOU performance break-down on UNC. }
\centering
\vspace{-0.2cm}
\begin{tabular}{c c c c c}
\toprule
\bf{Length} & \bf{1-2} & \bf{3} & \bf{4-5} & \bf{6-20} \\
\midrule
R + LSTM & 43.66 & 40.60 & 33.98 & 24.91 \\
R + RMI & 44.51 & 41.86 & 35.05 & 25.95 \\
Relative Gain & 1.94\% & 3.10\% & 3.15\% & 4.19\% \\
\bottomrule
\end{tabular}
\vspace{-0.2cm}
\label{tab: unc-iou}
\end{table}

\begin{table}[t]
\caption{IOU performance break-down on UNC+. }
\centering
\vspace{-0.2cm}
\begin{tabular}{c c c c c}
\toprule
\bf{Length} & \bf{1-2} & \bf{3} & \bf{4-5} & \bf{6-20} \\
\midrule
R + LSTM & 34.40 & 24.04 & 19.31 & 12.30 \\
R + RMI & 35.72 & 25.41 & 21.73 & 14.37 \\
Relative Gain & 3.84\% & 5.67\% & 12.55\% & 16.85\% \\
\bottomrule
\end{tabular}
\vspace{-0.2cm}
\label{tab: unc+-iou}
\end{table}

\begin{table}[t]
\caption{IOU performance break-down on ReferItGame. }
\centering
\vspace{-0.2cm}
\begin{tabular}{c c c c c}
\toprule
\bf{Length} & \bf{1} & \bf{2} & \bf{3-4} & \bf{5-20} \\
\midrule
R + LSTM & 67.64 & 52.26 & 44.87 & 33.81 \\
R + RMI & 68.11 & 52.73 & 45.69 & 34.53 \\
Relative Gain & 0.69\% & 0.90\% & 1.82\% & 2.10\% \\
\bottomrule
\end{tabular}
\vspace{-0.2cm}
\label{tab: referit-iou}
\end{table}

\begin{figure}[t]
\captionsetup[subfigure]{labelformat=empty, font=small}
\centering
\begin{subfigure}{0.46\textwidth}
\centering
\begin{subfigure}{0.24\textwidth}
\caption{\scriptsize{\textit{$\hdots$girl}}}
\vspace{-2mm}
\includegraphics[width=\linewidth]{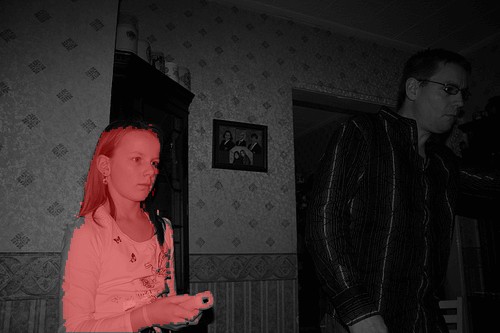}
\includegraphics[width=\linewidth]{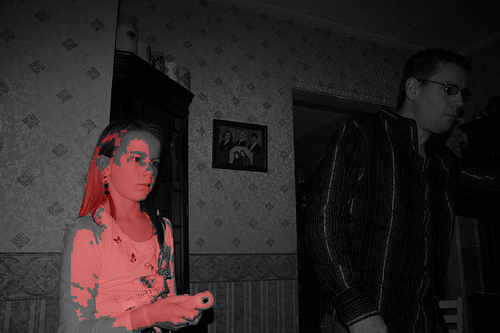}
\end{subfigure}
\begin{subfigure}{0.24\textwidth}
\caption{\scriptsize{\textit{$\hdots$white}}}
\vspace{-2mm}
\includegraphics[width=\linewidth]{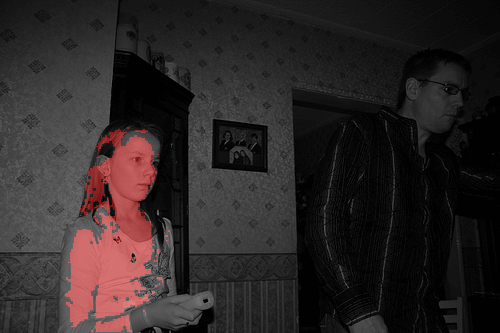}
\includegraphics[width=\linewidth]{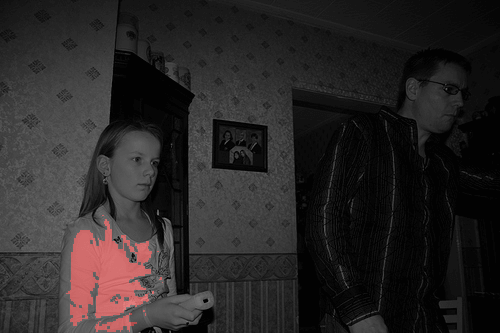}
\end{subfigure}
\begin{subfigure}{0.24\textwidth}
\caption{\scriptsize{\textit{$\hdots$Wii}}}
\vspace{-2mm}
\includegraphics[width=\linewidth]{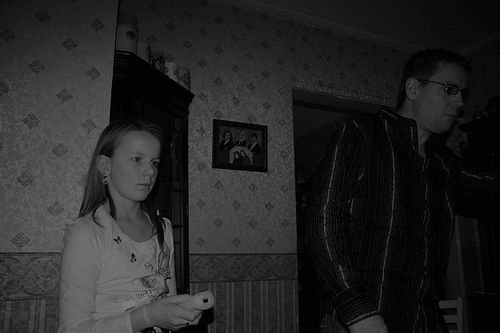}
\includegraphics[width=\linewidth]{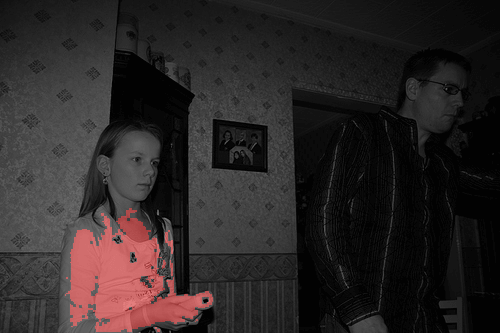}
\end{subfigure}
\begin{subfigure}{0.24\textwidth}
\caption{\scriptsize{\textit{$\hdots$remote}}}
\vspace{-2mm}
\includegraphics[width=\linewidth]{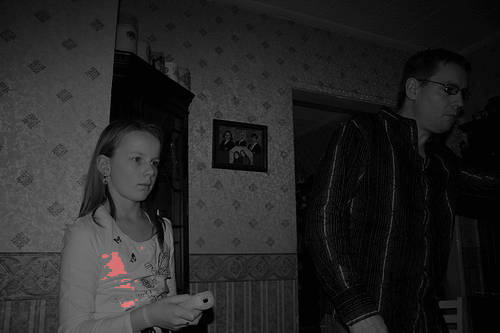}
\includegraphics[width=\linewidth]{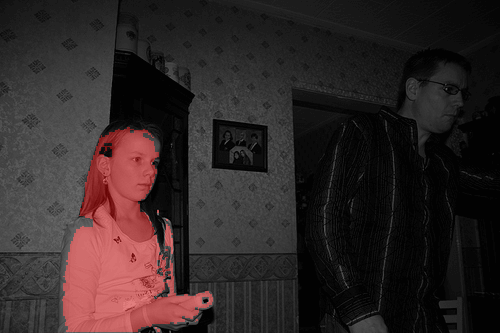}
\end{subfigure}
\caption{\textit{A \underline{girl} in \underline{white} holding a \underline{Wii} \underline{remote}.}}
\end{subfigure}

\begin{subfigure}{0.46\textwidth}
\centering
\begin{subfigure}{0.24\textwidth}
\caption{\scriptsize{\textit{$\hdots$train}}}
\vspace{-2mm}
\includegraphics[width=\linewidth]{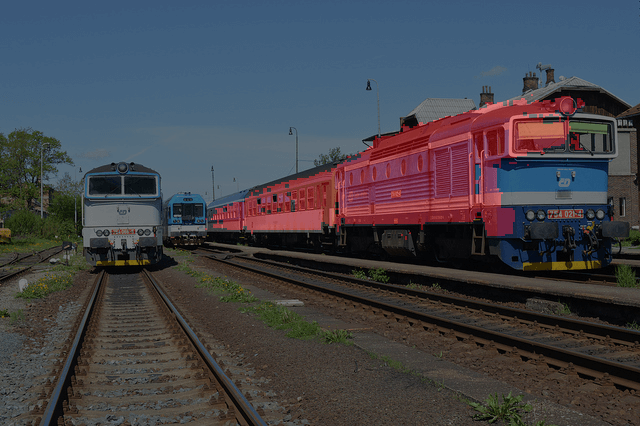}
\includegraphics[width=\linewidth]{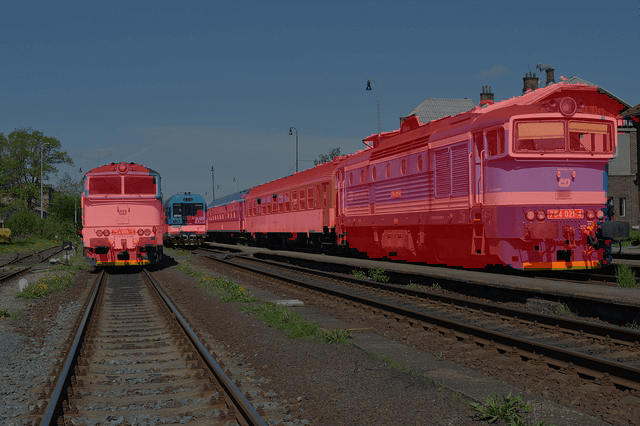}
\end{subfigure}
\begin{subfigure}{0.24\textwidth}
\caption{\scriptsize{\textit{$\hdots$right}}}
\vspace{-2mm}
\includegraphics[width=\linewidth]{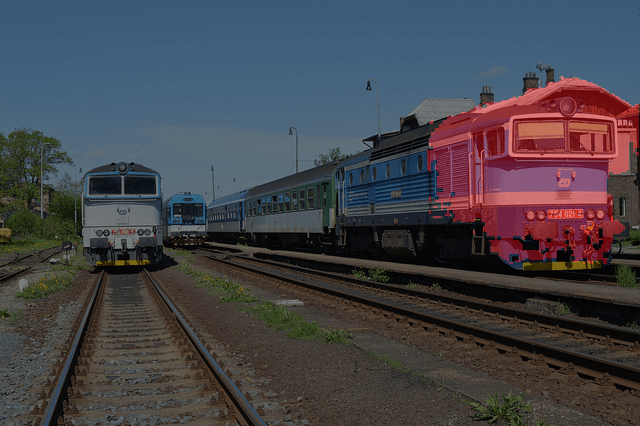}
\includegraphics[width=\linewidth]{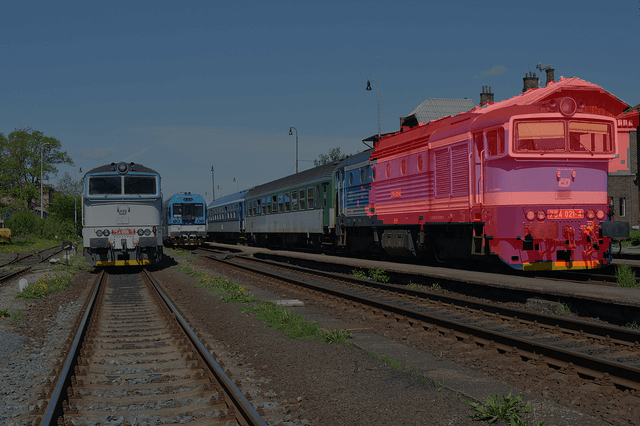}
\end{subfigure}
\begin{subfigure}{0.24\textwidth}
\caption{\scriptsize{\textit{$\hdots$driving}}}
\vspace{-2mm}
\includegraphics[width=\linewidth]{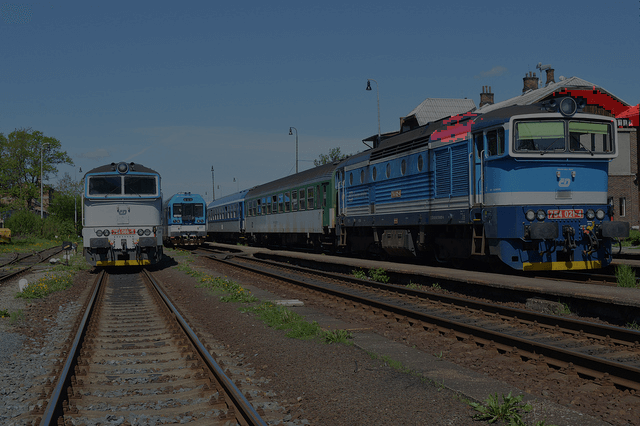}
\includegraphics[width=\linewidth]{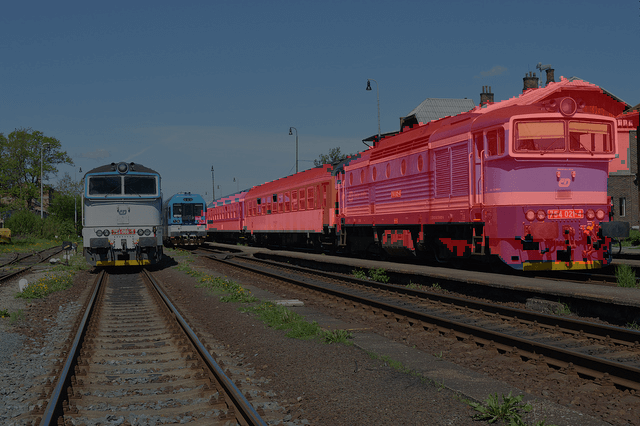}
\end{subfigure}
\begin{subfigure}{0.24\textwidth}
\caption{\scriptsize{\textit{$\hdots$trains}}}
\vspace{-2mm}
\includegraphics[width=\linewidth]{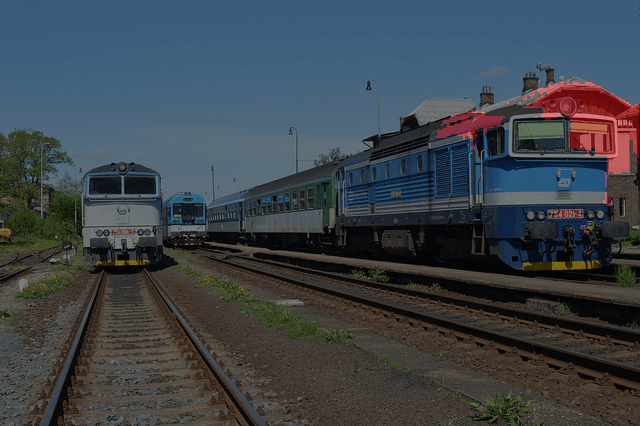}
\includegraphics[width=\linewidth]{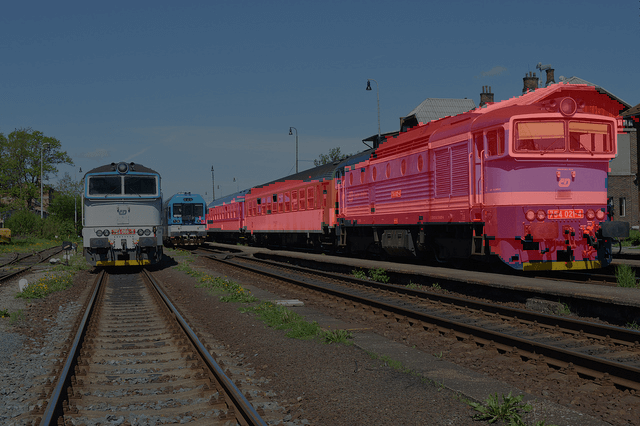}
\end{subfigure}
\caption{\textit{Blue \underline{train} on the far \underline{right} trail \underline{driving} ahead of two other \underline{trains}.}}
\end{subfigure}


\begin{subfigure}{0.46\textwidth}
\centering
\begin{subfigure}{0.24\textwidth}
\caption{\scriptsize{\textit{Dog}}}
\vspace{-2mm}
\includegraphics[width=\linewidth]{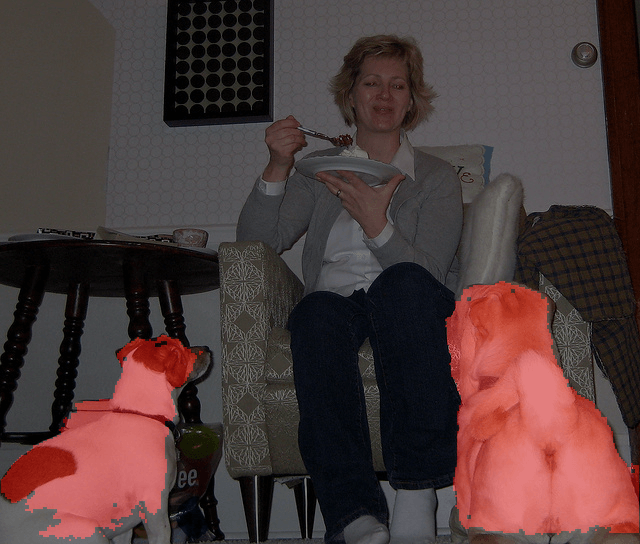}
\includegraphics[width=\linewidth]{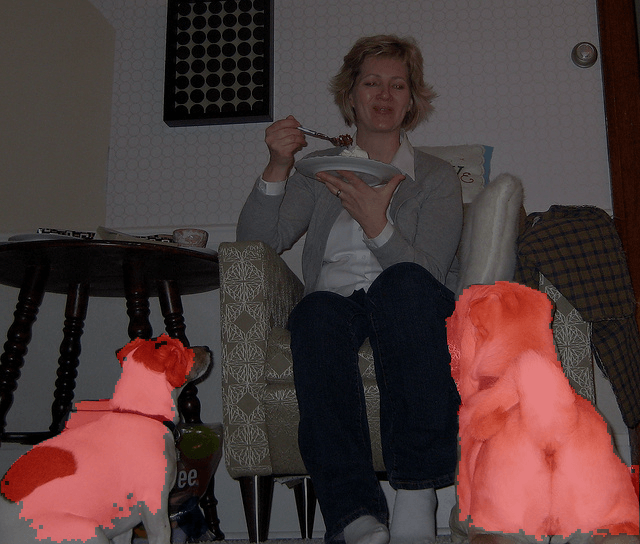}
\end{subfigure}
\begin{subfigure}{0.24\textwidth}
\caption{\scriptsize{\textit{$\hdots$close}}}
\vspace{-2mm}
\includegraphics[width=\linewidth]{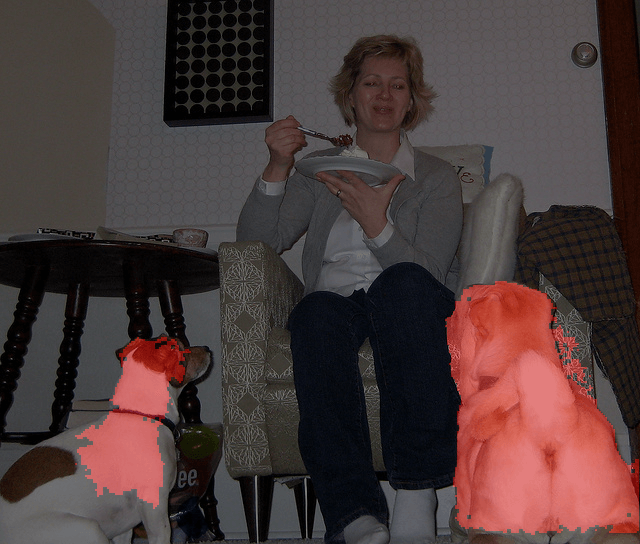}
\includegraphics[width=\linewidth]{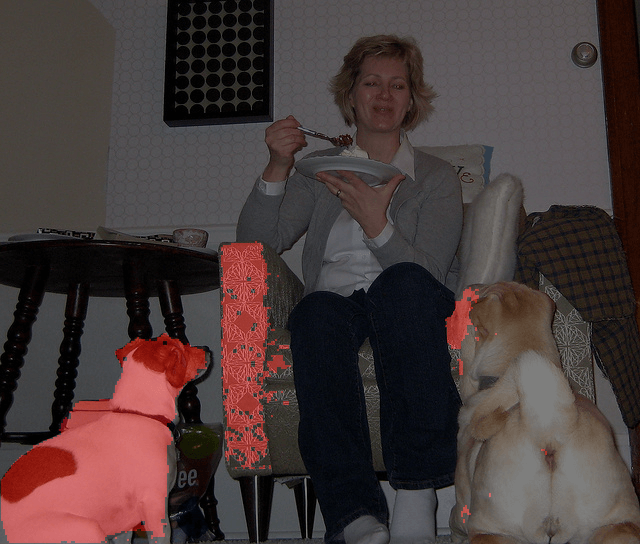}
\end{subfigure}
\begin{subfigure}{0.24\textwidth}
\caption{\scriptsize{\textit{$\hdots$tall}}}
\vspace{-2mm}
\includegraphics[width=\linewidth]{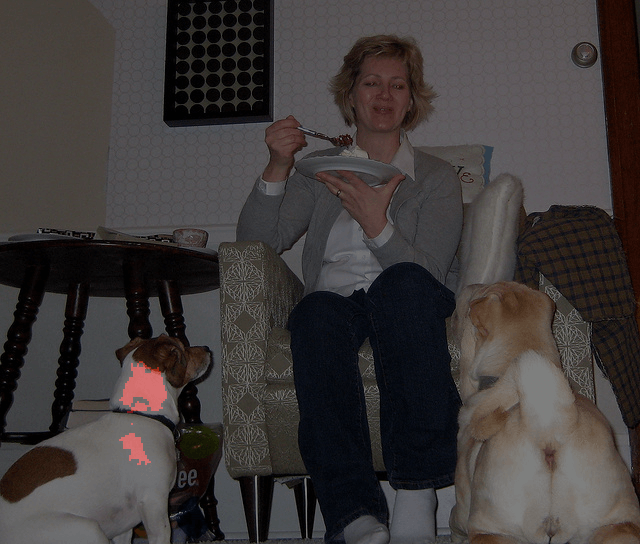}
\includegraphics[width=\linewidth]{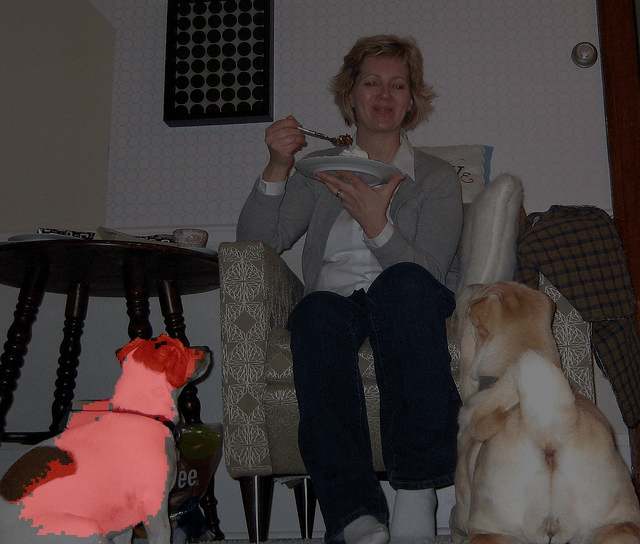}
\end{subfigure}
\begin{subfigure}{0.24\textwidth}
\caption{\scriptsize{\textit{$\hdots$table}}}
\vspace{-2mm}
\includegraphics[width=\linewidth]{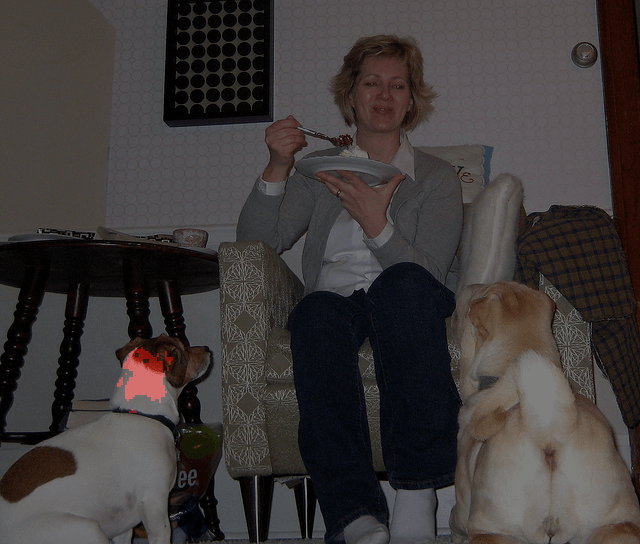}
\includegraphics[width=\linewidth]{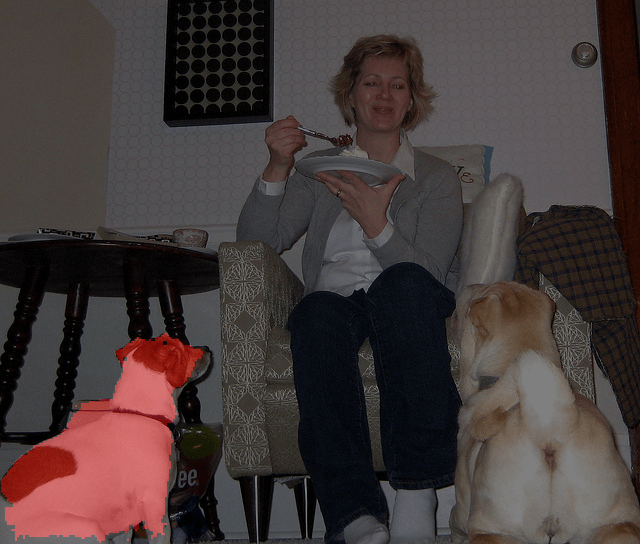}
\end{subfigure}
\caption{\textit{\underline{Dog} \underline{close} to the \underline{tall} \underline{table}.}}
\end{subfigure}


\caption{Comparison of D+LSTM+DCRF (first row) and D+RMI+DCRF (second row). Each column shows segmentation result until after reading the underlined word.}
\label{fig: long}
\vspace{-0.2cm}
\end{figure}

\begin{figure*}[t]
\captionsetup[subfigure]{labelformat=empty, font=small}
\centering
\begin{subfigure}{\textwidth}
\centering
\includegraphics[width=0.15\linewidth]{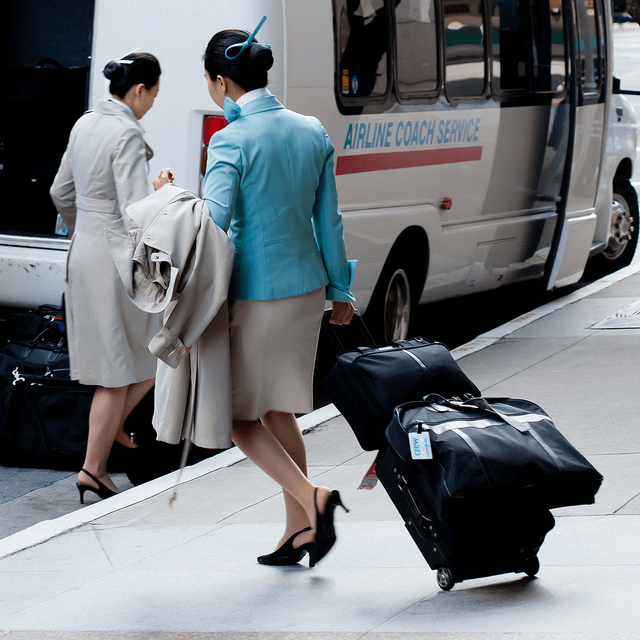}
\includegraphics[width=0.15\linewidth]{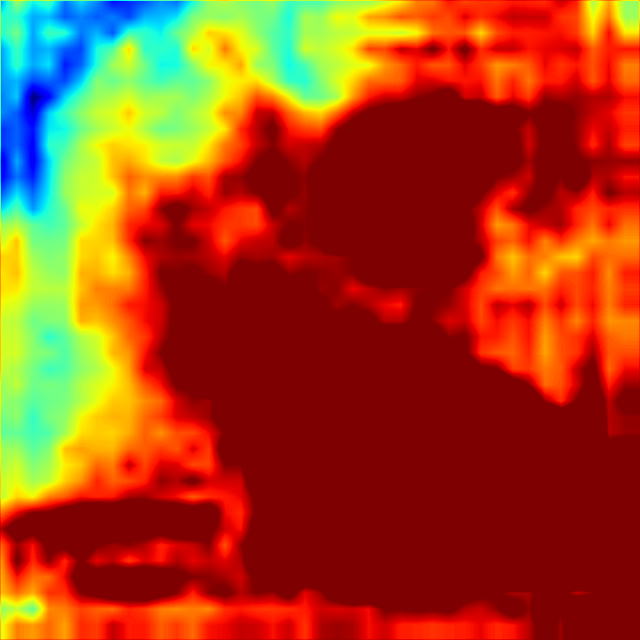}
\includegraphics[width=0.15\linewidth]{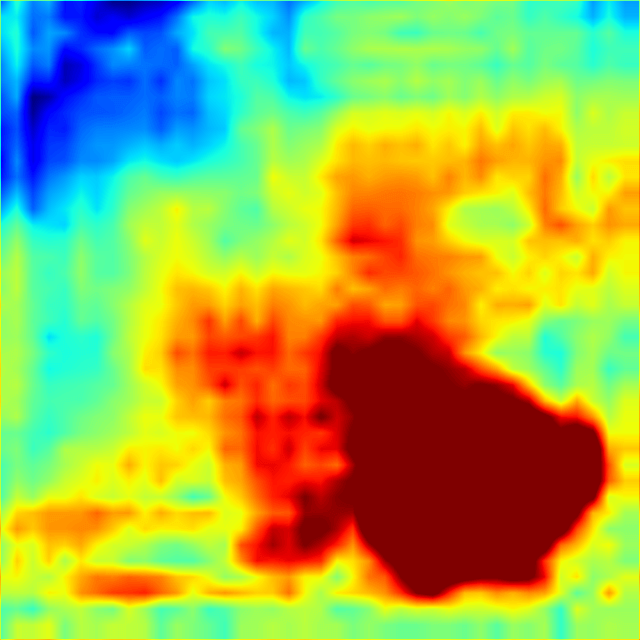}
\includegraphics[width=0.15\linewidth]{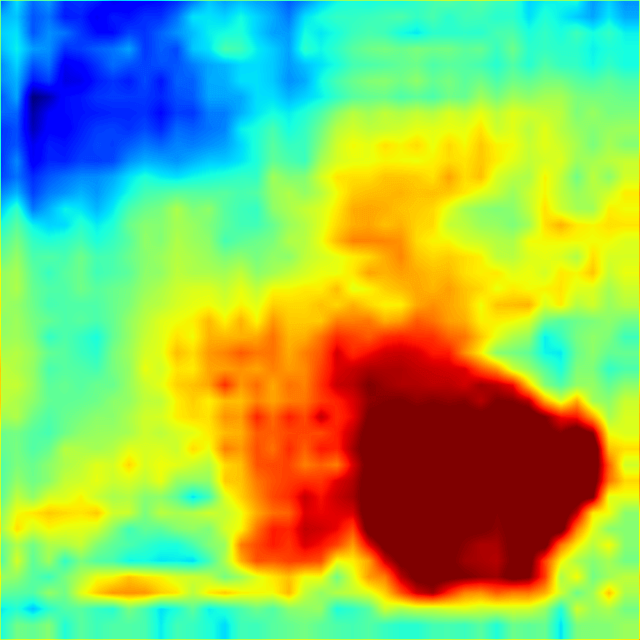}
\includegraphics[width=0.15\linewidth]{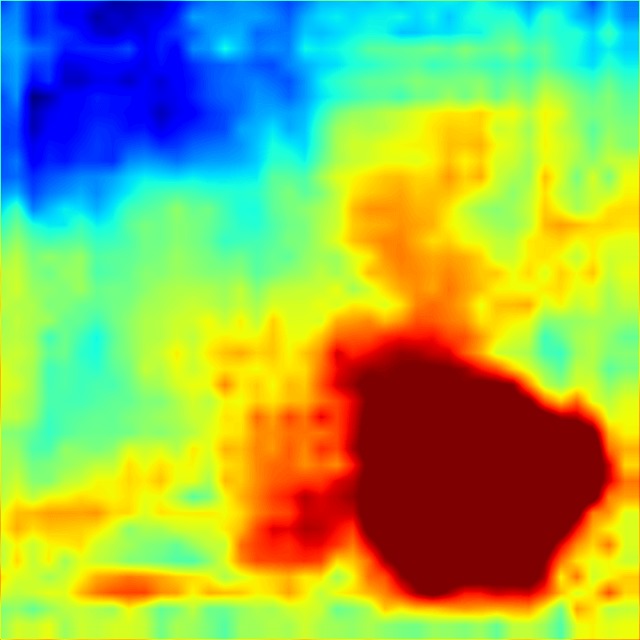}
\includegraphics[width=0.15\linewidth]{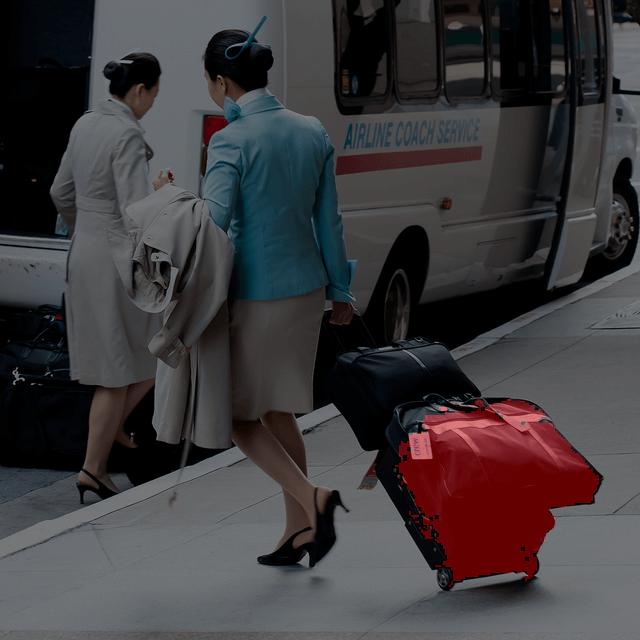}
\caption{\textit{The \underline{bottom} two \underline{luggage} \underline{cases} being \underline{rolled}.}}
\end{subfigure}

\begin{subfigure}{\textwidth}
\centering
\includegraphics[width=0.15\linewidth]{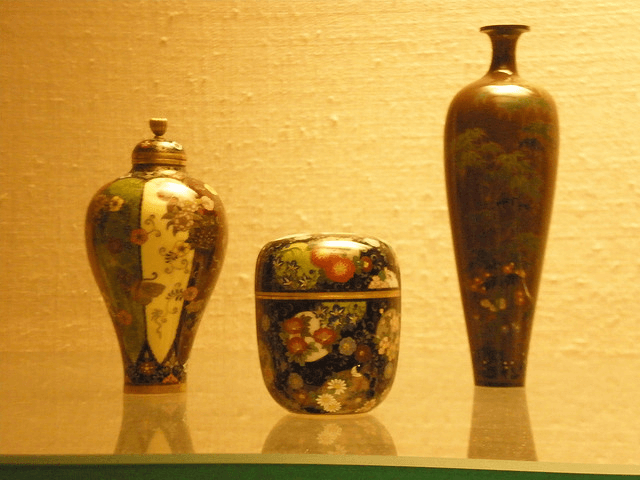}
\includegraphics[width=0.15\linewidth]{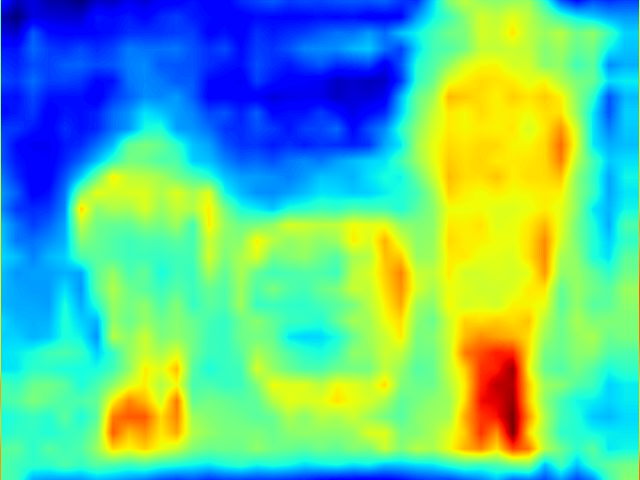}
\includegraphics[width=0.15\linewidth]{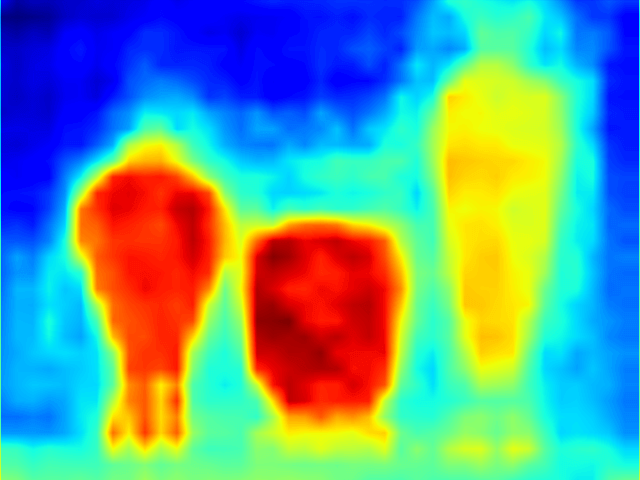}
\includegraphics[width=0.15\linewidth]{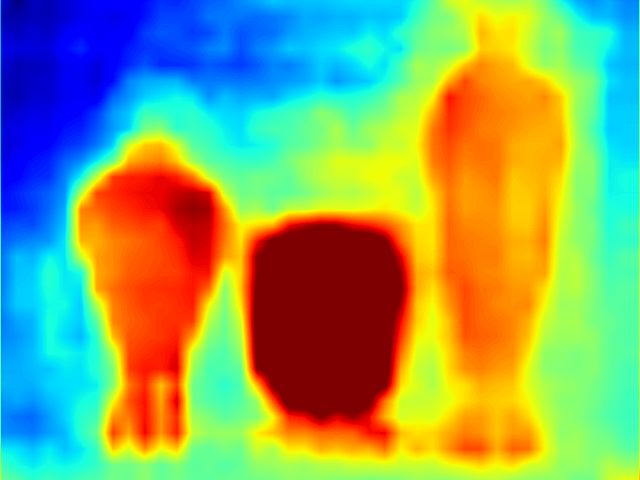}
\includegraphics[width=0.15\linewidth]{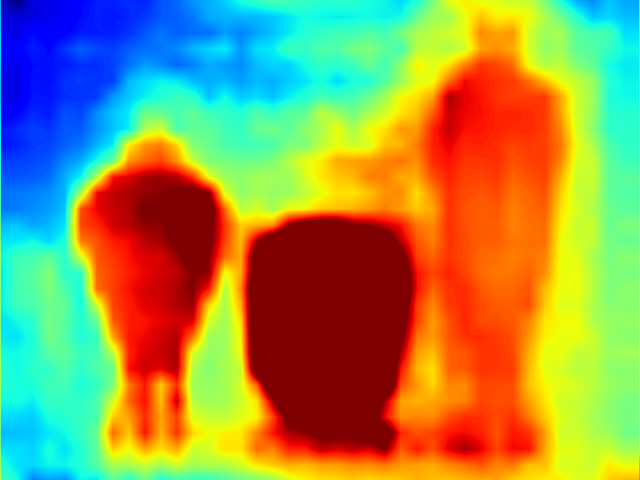}
\includegraphics[width=0.15\linewidth]{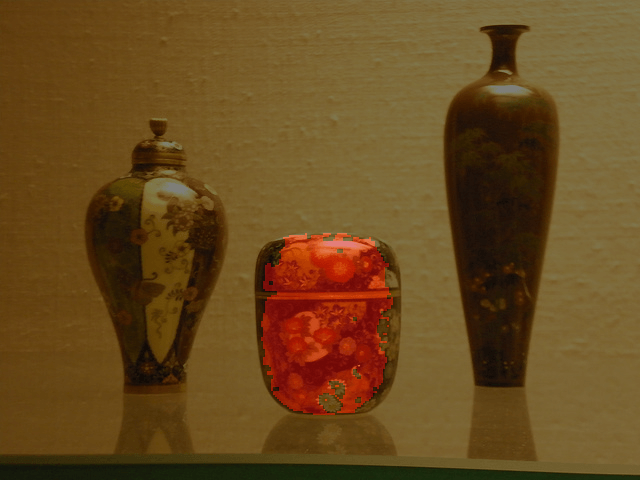}
\caption{\textit{\underline{The} small \underline{vase} in the \underline{middle} of the other \underline{vases}.}}
\end{subfigure}

\begin{subfigure}{\textwidth}
\centering
\includegraphics[width=0.15\linewidth]{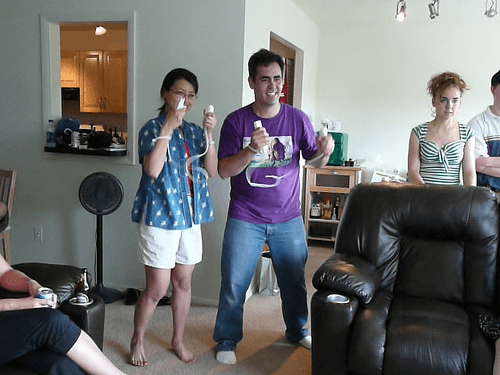}
\includegraphics[width=0.15\linewidth]{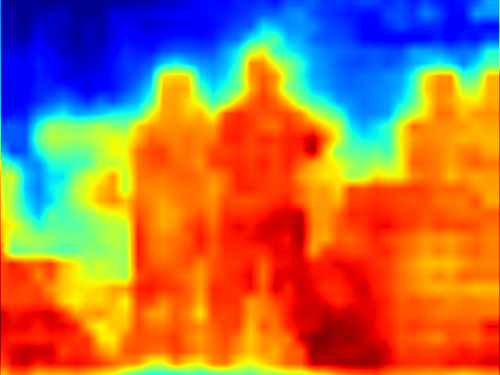}
\includegraphics[width=0.15\linewidth]{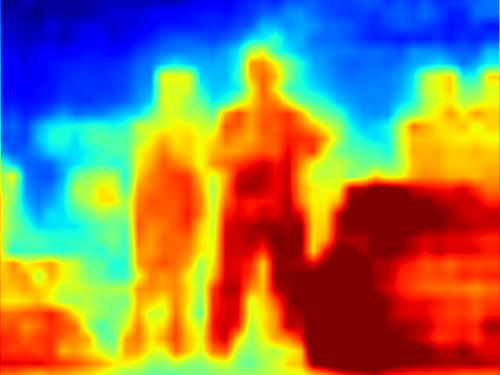}
\includegraphics[width=0.15\linewidth]{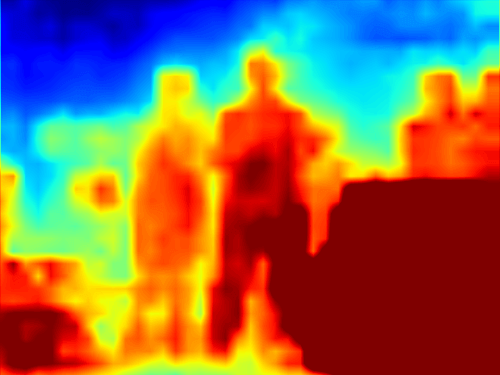}
\includegraphics[width=0.15\linewidth]{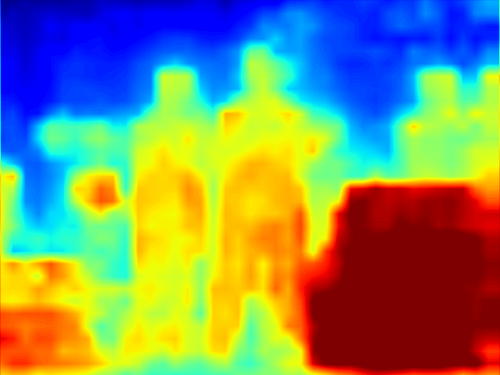}
\includegraphics[width=0.15\linewidth]{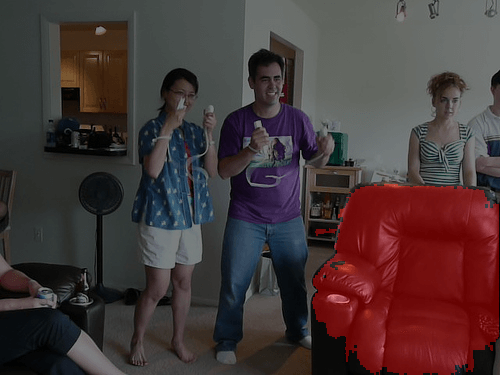}
\caption{\textit{\underline{An} empty \underline{leather} \underline{chair} with a cup holder built \underline{in}.}}
\end{subfigure}
\vspace{-0.2cm}
\caption{Visualizing and understanding convolutional multimodal LSTM in our RMI model. The first column is the original image, and the last column is the final segmentation output of D+RMI+DCRF. The middle columns visualize the output of mLSTM at underlined words by meanpooling the 500-dimensional feature. }
\label{fig: seq}
\vspace{-0.1cm}
\end{figure*}

\begin{figure*}[t!]
\captionsetup[subfigure]{labelformat=empty, font=small}
\centering
\begin{subfigure}{.15\textwidth}
\caption{Image}
\vspace{-2mm}
\end{subfigure}
\begin{subfigure}{.15\textwidth}
\caption{GT}
\vspace{-2mm}
\end{subfigure}
\begin{subfigure}{.15\textwidth}
\caption{D+LSTM}
\vspace{-2mm}
\end{subfigure}
\begin{subfigure}{.15\textwidth}
\caption{D+LSTM+DCRF}
\vspace{-2mm}
\end{subfigure}
\begin{subfigure}{.15\textwidth}
\caption{D+RMI}
\vspace{-2mm}
\end{subfigure}
\begin{subfigure}{.15\textwidth}
\caption{D+RMI+DCRF}
\vspace{-2mm}
\end{subfigure}
\begin{subfigure}{\textwidth}
\centering
\includegraphics[width=0.15\linewidth]{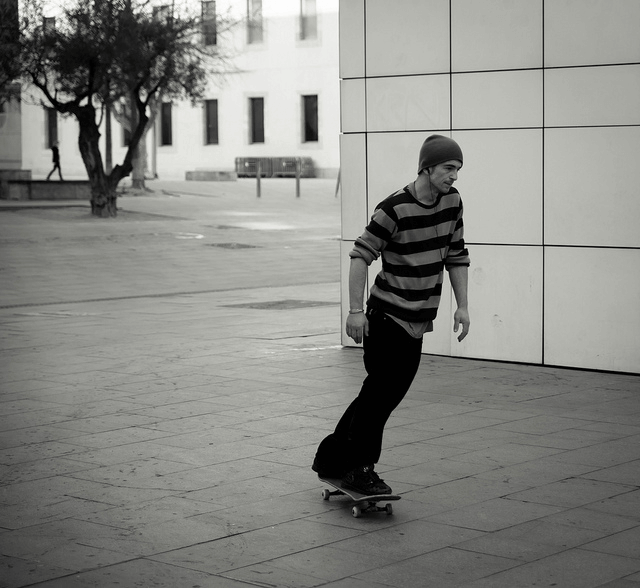}
\includegraphics[width=0.15\linewidth]{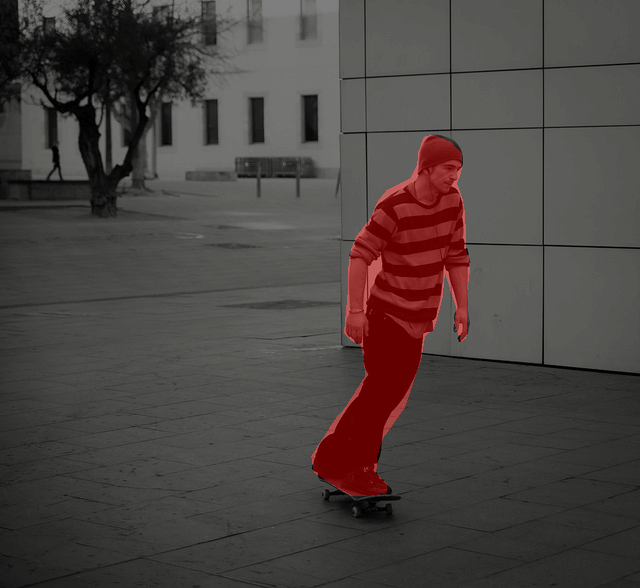}
\includegraphics[width=0.15\linewidth]{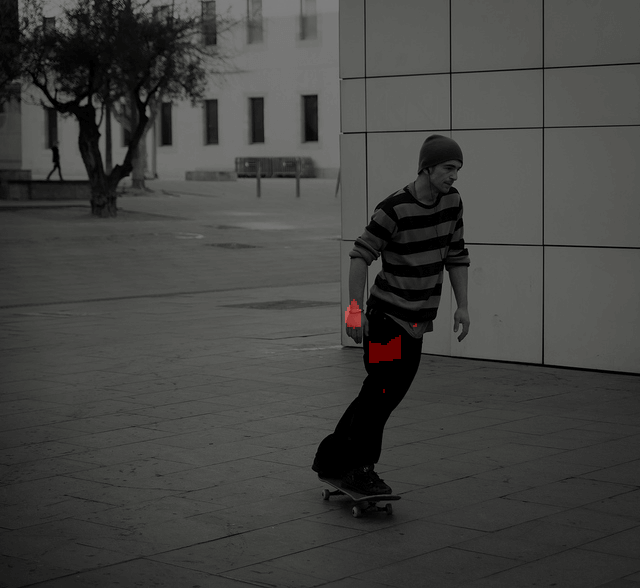}
\includegraphics[width=0.15\linewidth]{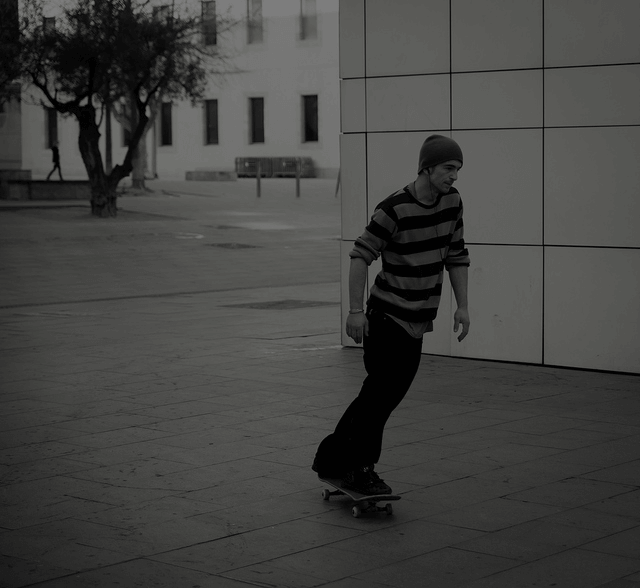}
\includegraphics[width=0.15\linewidth]{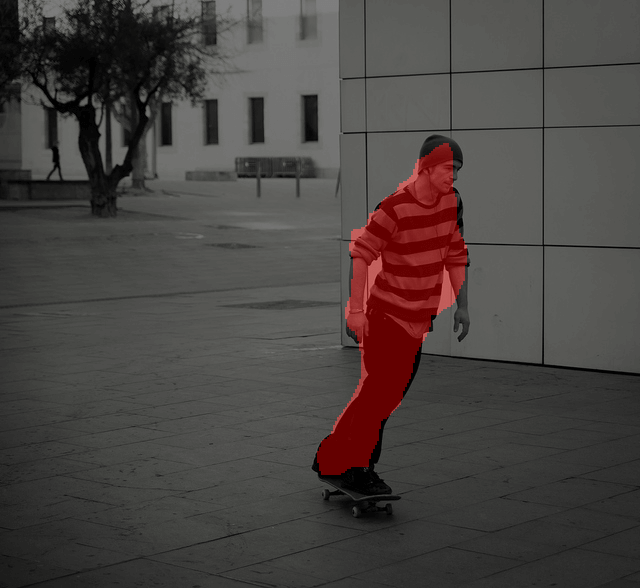}
\includegraphics[width=0.15\linewidth]{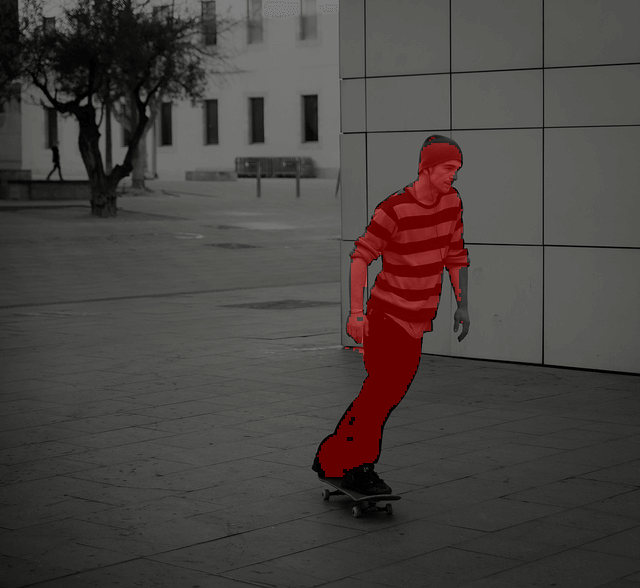}
\caption{\textit{A skateboarder skateboarding in a city listening to his music while turning around a corner.}}
\end{subfigure}
\begin{subfigure}{\textwidth}
\centering
\includegraphics[width=0.15\linewidth]{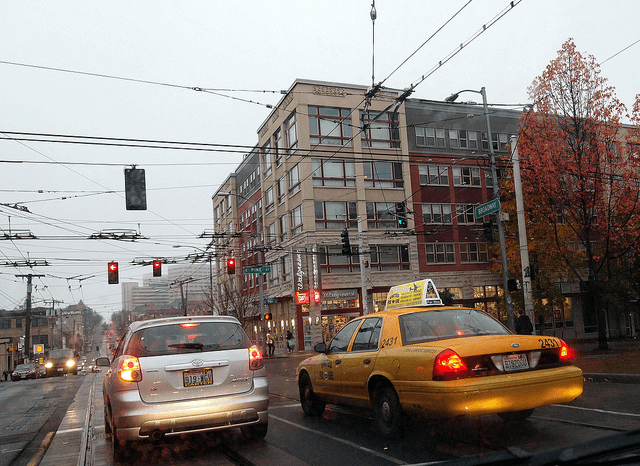}
\includegraphics[width=0.15\linewidth]{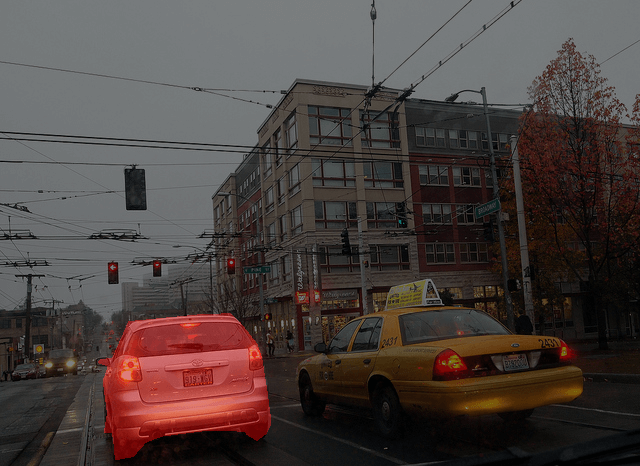}
\includegraphics[width=0.15\linewidth]{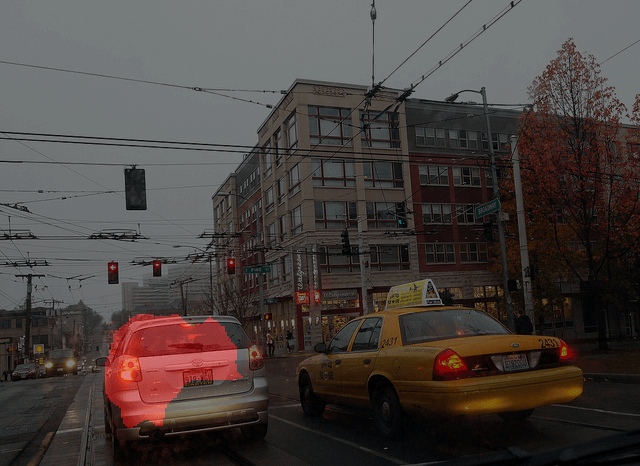}
\includegraphics[width=0.15\linewidth]{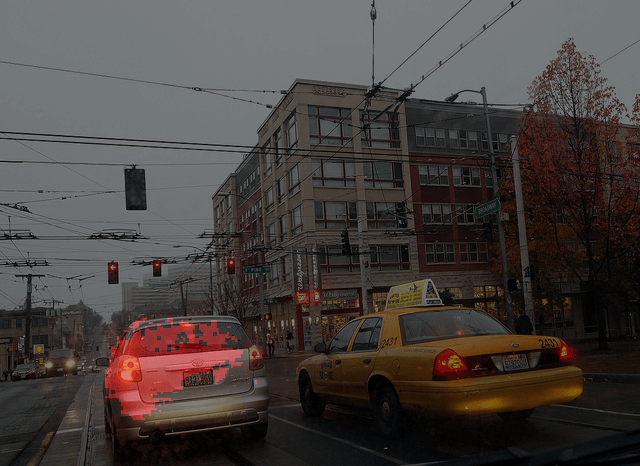}
\includegraphics[width=0.15\linewidth]{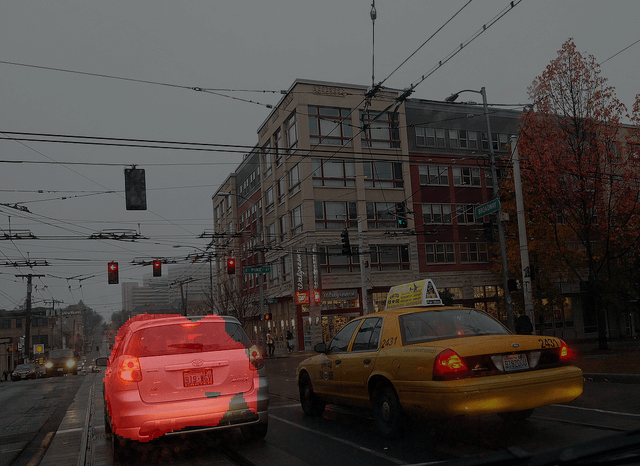}
\includegraphics[width=0.15\linewidth]{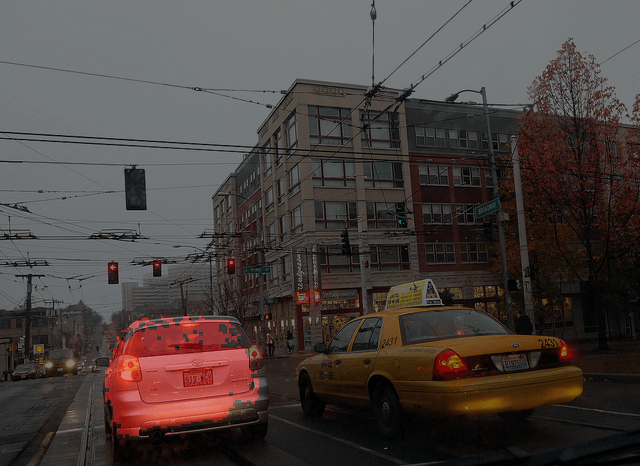}
\caption{\textit{Silver car on left.}}
\end{subfigure}
\begin{subfigure}{\textwidth}
\centering
\includegraphics[width=0.15\linewidth]{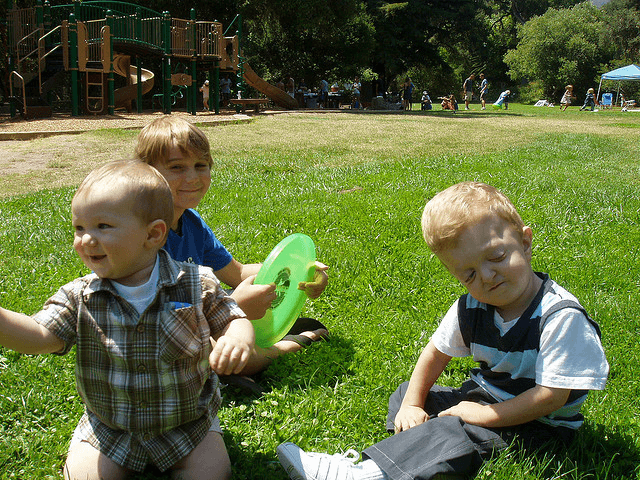}
\includegraphics[width=0.15\linewidth]{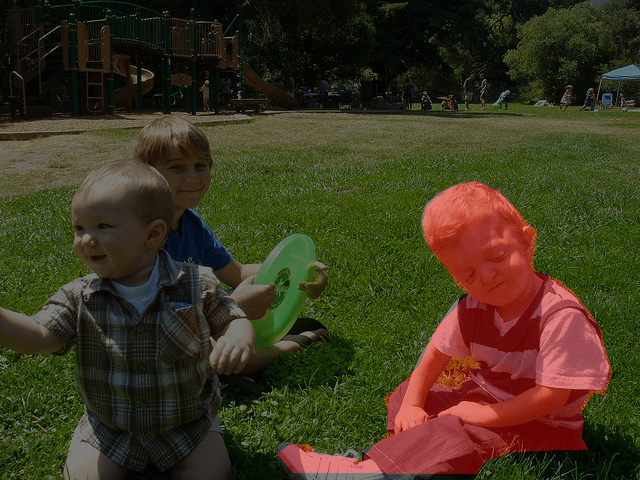}
\includegraphics[width=0.15\linewidth]{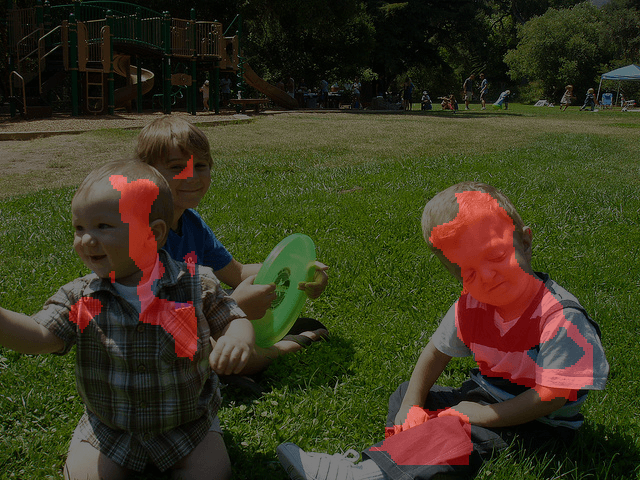}
\includegraphics[width=0.15\linewidth]{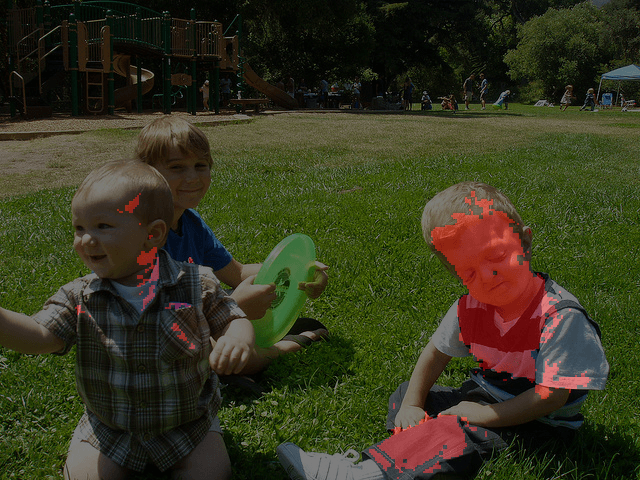}
\includegraphics[width=0.15\linewidth]{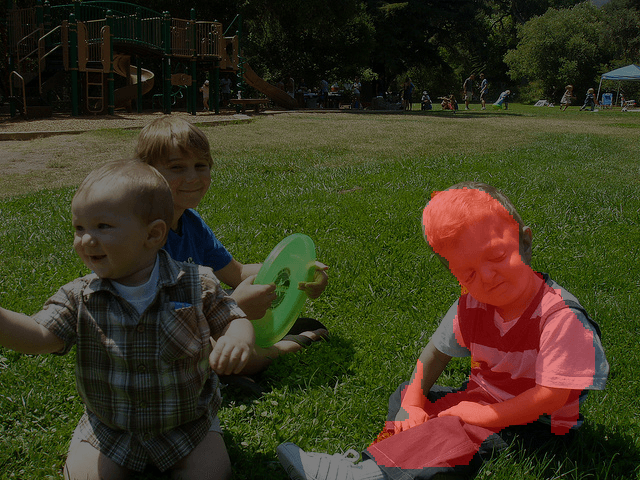}
\includegraphics[width=0.15\linewidth]{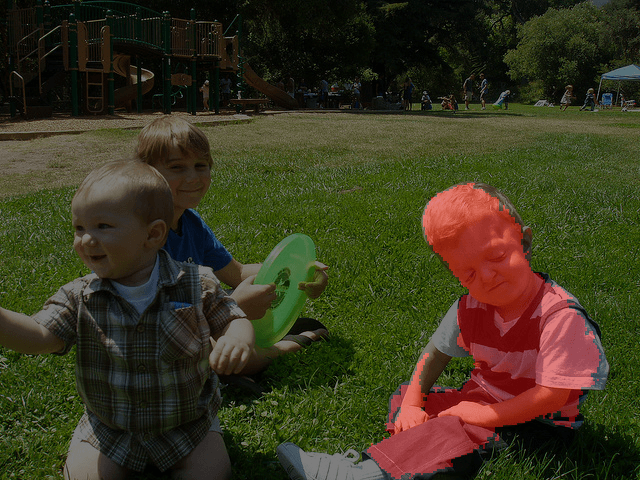}
\caption{\textit{Strip shirt boy eyes closed.}}
\end{subfigure}
\begin{subfigure}{\textwidth}
\centering
\includegraphics[width=0.15\linewidth]{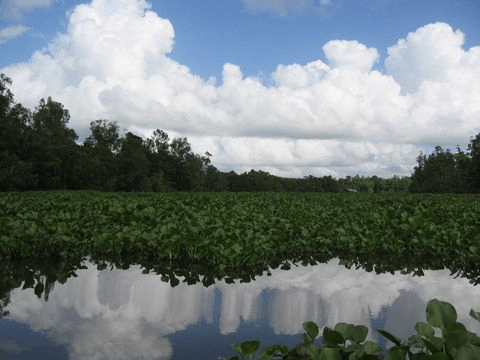}
\includegraphics[width=0.15\linewidth]{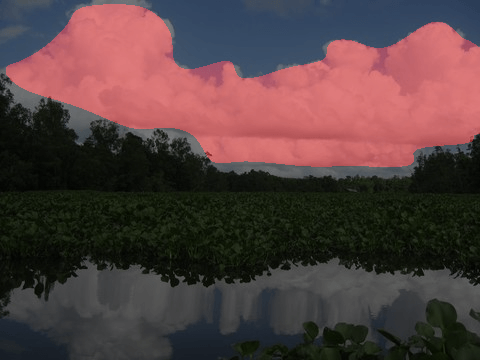}
\includegraphics[width=0.15\linewidth]{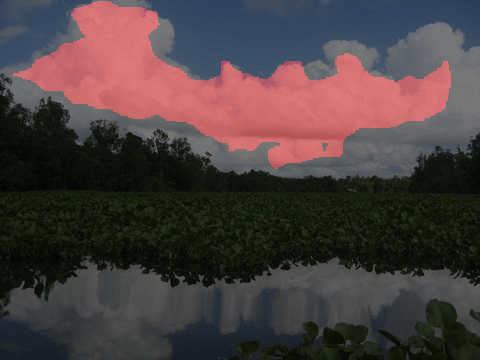}
\includegraphics[width=0.15\linewidth]{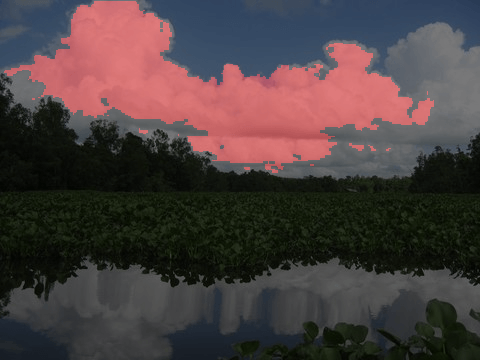}
\includegraphics[width=0.15\linewidth]{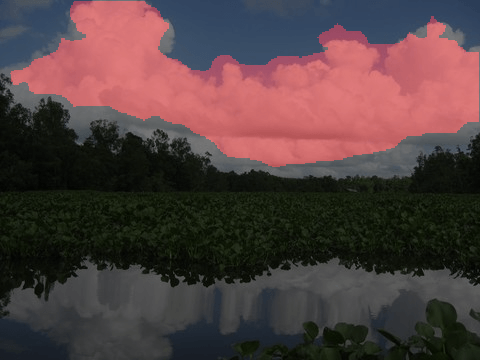}
\includegraphics[width=0.15\linewidth]{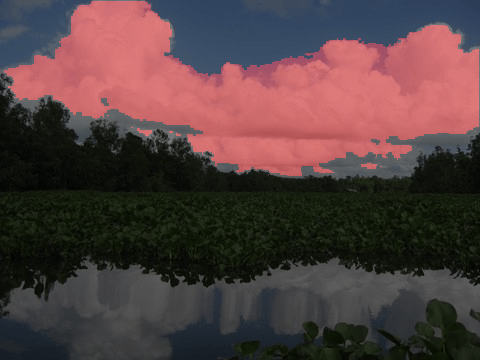}
\caption{\textit{Giant cloud.}}
\end{subfigure}
\vspace{-0.2cm}
\caption{Qualitative results of referring image segmentation. From top down are images from Google-Ref, UNC, UNC+, ReferItGame respectively. 
}
\label{fig: res}
\vspace{-0.2cm}
\end{figure*}

\subsection{Qualitative Results}

As aforementioned, our RMI model is better than the baseline model in modeling long sequences. 
We can see from the examples in Fig. \ref{fig: long} that the language-only LSTM is more easily distracted by words at the end of the sentence, resulting in unsatisfactory segmentation, while our model remains unaffected.

We suspect the reason is because our model can turn segmentation into a sequential process, saving the burden on the LSTM hidden state to encode the entire sentence. 
We are therefore interested in visualizing how the multimodal LSTM hidden state progresses over time.
Each mLSTM hidden state is a feature tensor of size $H' \times W' \times 500$.
We visualize this tensor by first doing bilinear interpolation and then collapsing the feature dimension via meanpooling, generating a $H \times W$ response map. 
We provide three examples in Fig. \ref{fig: seq}.
In the first example, after reading only "The bottom", the model is not sure about what objects is to be referred, and pays general attention to the bottom half of the image.
As soon as it reads "luggage", the response map pinpoints the objects, and remembers the information until the end of the sentence to generate the correct segmentation output.
In the second example, in the beginning after reading "The", the model appears unsure.
After reading "The small vase", it discards the largest vase and focuses on the other two.
As soon as the language mentions "middle", the response in the middle is enhanced, and retained till the end of the expression.
In the third example there is no location words, but the response around the correct region gradually enhances with "leather" and "chair", and the response on people is gradually suppressed after reading more words.
We can see that the mLSTM is successful at learning meaningful multimodal feature interaction in a sequential fashion that is consistent with our intuition in the introduction section. 
The meaningful multimodal features make it easier for the last convolution layer to do binary segmentation.

In Fig. \ref{fig: res} we provide some qualitative results of referring image segmentation on the four datasets.
For Google-Ref, the language understanding is more challenging.
In addition to handling longer sequences, it also needs to cope with all kinds of high level reasoning, e.g. "turning around a corner", and potentially redundant information, e.g. "listening to his music". 
For UNC, the expression is much shorter, and spatial words are allowed, e.g. "on left". 
For UNC+, the expression is more challenging. 
The image region could have just been described as "boy on right", but instead the model needs to reason from attributes like "strip shirt" and "eyes closed".
For ReferItGame, the segmentation target is more flexible as it contains "stuff" segments in addition to objects.
We show that by propagating the multimodal feature, our RMI model can better keep the intermediate belief, usually resulting in a more complete segmentation result.
The effect of DenseCRF is also clearly demonstrated.
For example, for the first image, DenseCRF can better refine the D+RMI result to align the prediction to the edges, and for the third image, DenseCRF can suppress the scattered wrong prediction in the D+LSTM result.

%% file: conc.tex
\section{Conclusion}

In this work we study the challenging problem of referring image segmentation. 
Learning a good multimodal representation is essential in this problem, since segmentation represents the correspondence or consistency between images and language.
Unlike previous work, which encodes the referring expression and image into vector representation independently, we build on the observation that referring image segmentation is a sequential process, and perform multimodal feature fusion after seeing every word in the referring expression.
To this end we propose the Recurrent Multimodal Interaction model, a novel two-layer recurrent architecture that encodes the sequential interactions between individual words, visual information, and spatial information as its hidden state.

We show the advantage of our word-to-image scheme over the sentence-to-image scheme.
Our model achieves the new state-of-the-art on all large-scale benchmark datasets.
In addition, we visualize the mLSTM hidden state and show that the learned multimodal feature is human-interpretable and facilitates segmentation.
In the future we plan to introduce more structure in language understanding.